%% file: main.tex
\documentclass[conference]{IEEEtran}

\usepackage[utf8]{inputenc} % allow utf-8 input
\usepackage[T1]{fontenc}    % use 8-bit T1 fonts
\usepackage{hyperref}       % hyperlinks
\usepackage{url}            % simple URL typesetting
\usepackage{booktabs}       % professional-quality tables
\usepackage{amsfonts}       % blackboard math symbols
\usepackage{nicefrac}       % compact symbols for 1/2, 
\usepackage{comment}
\usepackage{microtype}      % microtypography
\usepackage{xcolor}         % colors
\usepackage{multirow} 
\usepackage{graphicx}
\usepackage{subcaption}
\usepackage[linesnumbered]{algorithm2e}
\usepackage{adjustbox}
\usepackage{amsmath}
\usepackage{bbm}
\usepackage{amsthm}
\newtheorem*{theorem}{\textbf{Theorem}}
\newtheorem{lemma}{\textbf{Lemma}}

\newtheorem{assumption2}{\textbf{Assumption}}
\newcommand{\Hquad}{\hspace{0.6em}} 
\newcommand{\defeq}{\overset{\text{\tiny def}}{=}}
\def\BibTeX{{\rm B\kern-.05em{\sc i\kern-.025em b}\kern-.08em
    T\kern-.1667em\lower.7ex\hbox{E}\kern-.125emX}}

\input{commands}

\begin{document}

\title{On the Robustness of Tabular Foundation Models: Test-Time Attacks and In-Context Defenses
}
\begin{comment}
\author{\IEEEauthorblockN{1\textsuperscript{st}Anonymous authors}
\and
\IEEEauthorblockN{~2\textsuperscript{nd} Anonymous authors}
\and
\IEEEauthorblockN{3\textsuperscript{rd}Anonymous authors}
\and
\IEEEauthorblockN{4\textsuperscript{th}Anonymous authors}
\and
\IEEEauthorblockN{5\textsuperscript{th}Anonymous authors}
\and
\IEEEauthorblockN{6\textsuperscript{th}Anonymous authors}
\and
\IEEEauthorblockN{7\textsuperscript{th}Anonymous authors}}
\end{comment}
 \author{\IEEEauthorblockN{1\textsuperscript{st} Mohamed Djilani}
 \IEEEauthorblockA{\textit{SnT} \\
 \textit{University of Luxembourg}\\
 Luxembourg \\
 mohamed.djilani@uni.lu\\}
 \and
 \IEEEauthorblockN{2\textsuperscript{nd} Thibault Simonetto}
 \IEEEauthorblockA{\textit{SnT} \\
 \textit{University of Luxembourg}\\
 Luxembourg \\
 thibault.simonetto@uni.lu\\}
 \and
 \IEEEauthorblockN{3\textsuperscript{rd} Karim Tit}
 \IEEEauthorblockA{\textit{SnT} \\
 \textit{University of Luxembourg}\\
 Luxembourg \\
 karim.tit@uni.lu\\}
 \and
 \IEEEauthorblockN{4\textsuperscript{th} Florian Tambon}
 \IEEEauthorblockA{\textit{SnT} \\
 \textit{University of Luxembourg}\\
 Luxembourg \\
 florian.tambon@uni.lu\\}
 \and

 \IEEEauthorblockN{5\textsuperscript{th} Salah Ghamizi $^{\dagger}$}
 \IEEEauthorblockA{\textit{Luxembourg Institute of Health} \\
 \textit{SnT / University of Luxembourg}\\
 Luxembourg \\
 salah.ghamizi@lih.lu\\}
 \and
 \IEEEauthorblockN{6\textsuperscript{th} Maxime Cordy}
 \IEEEauthorblockA{\textit{SnT} \\
 \textit{University of Luxembourg}\\
 Luxembourg \\
 maxime.cordy@uni.lu\\}
 \and
 \IEEEauthorblockN{7\textsuperscript{th} Mike Papadakis}
 \IEEEauthorblockA{\textit{SnT} \\
 \textit{University of Luxembourg}\\
 Luxembourg \\
 michail.papadakis@uni.lu}
 }

\maketitle

\begingroup
\renewcommand\thefootnote{$\dagger$}
\footnotetext{This work was conducted while at Luxembourg Institute of Health.}
\endgroup

\begin{abstract}
Recent tabular Foundational Models (FM) such as TabPFN and TabICL, leverage in-context learning to achieve strong performance without gradient updates or fine-tuning. However, their robustness to adversarial manipulation remains largely unexplored. In this work, we present a comprehensive study of the adversarial vulnerabilities of tabular FM, focusing on both their fragility to targeted test-time attacks and their potential misuse as adversarial tools. We show on three benchmarks in finance, cybersecurity and healthcare, that small, structured perturbations to test inputs can significantly degrade prediction accuracy, even when the training context remains fixed. 
Additionally, we demonstrate that tabular FM can be repurposed to generate transferable evasion to conventional models such as random forests and XGBoost, and on a lesser extent to deep tabular models. 
To improve tabular FM, we formulate the robustification problem as an optimisation of the weights (adversarial fine-tuning), or the context (adversarial in-context learning). We introduce an in-context adversarial training strategy that incrementally replaces the context with adversarial perturbed instances, without updating model weights. Our approach improves robustness across multiple tabular benchmarks. Together, these findings position tabular FM as both a target and a source of adversarial threats, highlighting the urgent need for robust training and evaluation practices in this emerging paradigm.
\end{abstract}

%\IEEEpeerreviewmaketitle

\input{./doc/01_introduction}

\input{./doc/02_background}

\input{./doc/02_related}

\input{./doc/02_problem}

\input{./doc/03_experimental_protocol}

\input{./doc/04_rq1}

\input{./doc/05_rq2}

\input{./doc/06_rq3}

\input{./doc/06_bis_discussion}

\input{./doc/07_limitations}

\input{./doc/08_conclusion}

\section*{LLM usage considerations}

LLMs were used for editorial purposes in this manuscript, and all outputs were inspected by the authors to ensure accuracy and originality. Similarly, LLMs were used to assist in writing code. All code was manually verified and tested to ensure coherence in the results.

\section*{Acknowledgements}
This research was funded in whole, or in part, by the Luxembourg National Research Fund (FNR), grant references NCER22/IS/16570468/NCER-FT, BRIDGES/2022/IS/17437536, CORE C23/IS/18182513/MiCE and 18886361. 
Dr. Papadakis is supported by the Luxembourg National Research Fund (FNR) INTER/MOBILITY/2024/IS/18956086. 
Dr. Ghamizi is supported by the Luxembourg National Research Fund (FNR) CORE C24/IS/18942843.
This research was also supported by BGL BNP Paribas Luxembourg and the European Union's Horizon Europe research and innovation programme [HORIZON-CL4-2024-DATA-01-01] (grant No. 101189650).

\clearpage

\bibliographystyle{IEEEtran}
\bibliography{used_ref}

% \clearpage
% \input{doc/response_to_reviewers}

% \clearpage
% \input{doc/09_cover_letter}

\clearpage
\appendix
\input{doc/80_appendix}

%\input{doc/90_checklist}

\end{document}

%% file: commands.tex
\newcommand{\tablespace}{\addlinespace[0.2cm]}

%% file: doc/01_introduction.tex
% \section{Intro rephrased}
\section{Introduction}

Tabular machine learning is the backbone of critical decision-making systems in domains like finance, healthcare, and cybersecurity. These systems generally require evidence of their accuracy, interpretability, and robustness before being deployed in production. While deep neural networks achieve state-of-the-art results in vision and language tasks, they often match but rarely outperform gradient-boosted decision trees (GBDT) such as XGB~\cite{XGBOOST} in tabular data settings, despite requiring orders of magnitude more computational resources and data~\cite{tabsurvey}.

Emerging Tabular Foundational Models (FMs) such as TabPFN and TabICL~\cite{hollmann2025tabpfn,qu2025tabicl} promise to bring fundamental changes in this area, enabling in-context prediction on tabular data through synthetic pre-training. For example, TabPFNv2 (the second release of the TabPFN models family) achieves 5000x speedups over GBDTs and competitive performance on various tasks~\cite{hollmann2025tabpfn}. While this indicates the potential for these models to hold their promises of broad applicability at low cost (both human and computational), it is essential to study the trustworthiness of these models beyond accuracy to enable faithful use. In particular, it is essential to study the security properties of tabular FMs as their different learning mechanisms (transformer-based architectures, reliance on synthetic data for pre-training, and in-context learning) may bring new attack surfaces not present in traditional methods. %In particular, the requirements of recent regulations in the US and EU (the AI Act \cite{EU_AI_reg} for instance) impose high requirements on large FM in terms of robustness and trustworthiness.

Recent studies have investigated different trustworthiness properties of TabPFNv2, including generalisation \cite{ye2025closerlooktabpfnv2} and robustness to distribution drift/out-of-distribution samples~\cite{helli2024driftresilienttabpfnincontextlearning,wu2025zeroshotmetalearningtabularprediction}. Conversely, other studies \cite{zhao2024universalvulnerabilitieslargelanguage} explored the security against backdoor attacks of non-tabular in-context learners. However, to the best of our knowledge, no work has explored the robustness of tabular FMs to security attacks, in particular evasion attacks.

Evasion attacks pose significant risks in critical tabular applications (e.g.\ financial fraud and cyber-threat detection), where adversaries can manipulate input features to bypass model decisions. A recent benchmark~\cite{simonetto2024tabularbench} proposed Constrained Adaptive Attack (CAA) to evaluate adversarial robustness specifically in tabular settings. The authors demonstrated the vulnerability of all deep tabular models to evasion attacks under realistic scenarios, considering domain constraints and tabular ML peculiarities (categorical features, relationships between features, etc.). However, this study did not consider tabular FMs, a new learning paradigm, and how attackers can exploit the genericity of tabular FMs to harm other models and systems. In contrast, another study \cite{anwar2024adversarialrobustnessincontextlearning} comprehensively investigates the robustness to evasion attacks of in-context learners, but they focus on linear transformers, excluding tabular FMs. 
Ultimately, this raises the question of whether the deployment of tabular FMs can lead to secure applications that are robust to evasion attacks.

In this paper, we aim to answer this question and challenge the relevance of tabular FMs in real-world industrial settings from a security standpoint. To do so, we consider three real-world benchmarks from~\cite{simonetto2024tabularbench}: URL (cyber-security), LCLD (finance) and WIDS (healthcare).
We first demonstrate the poor adversarial robustness of the mainstream tabular FMs, namely TabPFNv2 and TabICL. Specifically, the robust accuracy of these models against CAA on LCLD is 8.5\% and 10.2\% (respectively), which is significantly lower than the most robust specialised models (STG, 52.6\%). Additionally, we study transferability and reveal that (1) tabular FMs are also significantly vulnerable to transfer attacks and (2) they can be used as an easy-to-build surrogate, leading to competing (and sometimes stronger) transfer attacks than when using a manually crafted surrogate (i.e.\ a classical model).

Given the bleak picture we draw of tabular FMs, we study the possibility of hardening such models against adversarial attacks. Besides the classical adversarial fine-tuning (AFT) training formulation (optimising model weights)~\cite{madry2017towards} that we adapt to Tabular FMs specificites, we also explore the possibility of hardening the tabular FMs using their in-context learning capability. To do so, we defined the in-context learning problem as a min-max optimisation problem over said context that we name Adversarial In-Context Learning (AICL). We then evaluate the ability of AICL the classical AFT. Our results show that both mechanisms could improve the initial robustness of the FM (e.g.\ from 9.9\% up to 41.2\% for AICL on LCLD) and that AICL outperforms the classical AFT formulation. Moreover, AICL can prove to be more consistent than traditional Madry's defence applied on non-FM models, though it can not reach the robustness of adversarially trained classical models (e.g., STG, LCLD, 81.5\%). Still, these results show our new formulation is sound (as it improves robustness compared to the vanilla FM), which motivates further research on robust defences for tabular FMs. We provide a package to replicate our results on \url{https://figshare.com/projects/TabFM/249944}.

\paragraph{Our contribution can be summarised as follows}

\begin{itemize}
\item We evaluate the robustness to evasion attacks of the recent tabular FMs, an emerging paradigm for tabular machine learning. We show that these models are as vulnerable as tree- and deep-learning-based models, sometimes even more.
\item We evaluate the evasion attacks in transferability settings with traditional models and foundational models, both as sources and targets of such attacks. We demonstrate the vulnerabilities enabled by tabular FMs due to their low training data requirements and their potential use as surrogates.
\item We adapt the classical adversarial training defence method for tabular FMs, inspired by Madry adversarial training, leveraging finetuning and in-context learning to produce two defences: classical adversarial training (AFT) and in-context defence (AICL). We show that the classical formulation, AFT, does not benefit tabular FM much when AICL prove to be better, opening a new venue of research.
\end{itemize}

%In addition, their study does not consider realistic threat models, nor SoTA attack mechanisms such as CAA \cite{simonetto2024constrained} that combine gradient and search optimization to break weak defenses. 

%% file: doc/02_background.tex
\section{Prior Information}

\textcolor{black}{Tabular FMs are transformers that were pre-trained on large-scale synthetic tabular data to learn a universal representation of such data. During inference, leveraging in-context learning, which is using the provided context in their prompt, they can do zero-shot inference on any new task using only said provided context.} The first version of TabPFN~\cite{hollmann2022tabpfn} first extends an instance $x_i$ with padding to a fixed dimension (e.g., 100), then the features are projected to a higher dimension for further processing. The label $y_i$ is preprocessed in a similar way, and the embeddings are then processed in a transformer and then in a 10-class classifier. The training set is generated with structured causal models and Bayesian neural networks, and the best checkpoint is selected with validation on real-world datasets. This initial release was limited to small-scale tasks, given the computational challenges posed by training larger Transformers.

TabPFNv2~\cite{hollmann2025tabpfn} introduces a dedicated feature tokeniser, and a random position encoding allows for distinguishing the different features. Thanks to various improvements in the attention architecture, this new approach supports a much increased number of features ($d<500$) and context size ($N < 10000$). Since its inception, several improvements to TabPFN have been introduced, either to improve its scalability~\cite{liu2025tabpfnunleashedscalableeffective}, the effectiveness on downstream specialised tasks~\cite{ruiz2024tabpfn}, or generalization~\cite{qu2025tabicl}. Without loss of generalization, we focus our evaluation on TabICL~\cite{qu2025tabicl} and TabPFN2~\cite{hollmann2025tabpfn}, given that these two approaches show the best generalization capabilities. We defer the discussion of alternative approaches to the Appendix~\ref{sec:app-related-comp}.

However, these approaches do not consider the fundamental vulnerability to evasion attacks inherent to TabPFN's designs. Given FMs leverage large transformers, they can be more susceptible to gradient attacks. In particular, these models are based on complex feature encoders to handle heterogeneous types and inter-feature dependencies that would be suitable targets for imperceptible attacks. 
Another specific risk are the interactions between the context and the attention weights learned by the FM, which opens a new surface of attack that traditional robustification methods, such as adversarial training, could not effectively mitigate with only weights update. Given the potential new threats and vulnerabilities raised by large foundation models with in-context learning, a better understanding and assessment of their vulnerability to evasion attacks becomes critical before their widespread deployment in critical scenarios.

%% file: doc/02_related.tex
\section{\textcolor{black}{Related Works}}

\subsection{Tabular Adversarial Attacks}

Several approaches have been proposed to attack tabular data. Ballet et al. introduce LowProFool \cite{ballet2019imperceptible}, an adversarial attack tailored for tabular data that prioritizes imperceptibility by minimizing perturbations on features deemed important by domain experts. The method uses a feature importance-weighted norm to guide the attack, ensuring that changes are concentrated on less critical features. Ben-Tov et al. \cite{ben2024cafa} propose CaFA, a system for generating cost-aware, feasible adversarial attacks against neural tabular classifiers. CaFA uses TabPGD to create adversarial examples in feature space, then projects them onto mined database constraints (denial constraints) to ensure realizability. Experiments show CaFA achieves higher feasible success rates while perturbing fewer features and with lower magnitudes than prior work. Simonetto et al. \cite{simonetto2024constrained} introduce CAA, a powerful evasion attack combining the gradient-based CAPGD and the search-based MOEVA to assess the robustness of deep tabular models. CAPGD improves on prior gradient attacks by using adaptive step sizes and momentum, achieving up to higher success rates. CAA further enhances effectiveness, reducing model accuracy by up to while being up to five times faster than MOEVA alone. The attack supports feature constraints and is proposed as a new benchmark for evaluating defenses in tabular machine learning. Ultimately, we chose CAA and CAPGD in this study, as the attacks subsumed the other available at the time and come with curated datasets and constraints. CaFa is also a relevant candidate, yet it is not applicable when number of constraints is too high, because of its reliance with FastADC, which is the case in the datasets used in our experiments

\subsection{Adversarial Robustness in In-Context Learning}

While in-context learning (ICL) has been extensively studied in the context of natural language processing, its application to tabular data is a burgeoning area of research. Recent studies have begun to explore the adversarial robustness of ICL in tabular settings. \cite{anwar2024adversarialrobustnessincontextlearning} investigates the susceptibility of single-layer linear transformers, which emulate gradient descent in-context, to \emph{hijacking attacks}. They demonstrate that perturbing a single example in the in-context training set can coerce the model into producing arbitrary predictions. While these attacks are effective on linear transformers, they do not readily transfer to more complex architectures like GPT-2. However, such models can still be compromised using gradient-based adversarial techniques. The study also finds that adversarial training, even when applied solely during fine-tuning, enhances robustness against these attacks. Interestingly, training against weaker attack models can sometimes confer resilience to stronger adversarial strategies. Complementing this, Yu et al.~\cite{yu2024evaluating} examine the robustness of retrieval-augmented ICL methods. Their findings indicate that while retrieval mechanisms can bolster defence against test sample perturbations, they may inadvertently increase vulnerability to attacks targeting the demonstrations themselves. To address this, they propose DARD, a training-free adversarial defense that enriches the example pool with adversarially perturbed samples, leading to improved performance and robustness. In the realm of natural language processing, He et al.~\cite{he2024using} explore the integration of natural language explanations (NLEs) into ICL. Their approach involves augmenting prompts with human-generated NLEs, which the model then uses to generate further explanations. This method yields over a 6\% improvement in robustness across various adversarial datasets, including HANS, ISCS, and ANLI, demonstrating the efficacy of NLEs in enhancing model resilience. Do et al.~\cite{long-etal-2024-prompt} used a GAN approach to optimise the prompt as an ICL setting. They evaluate their method on 13 NLP tasks, notably MMLU and GSM8K, improving previous results. In contrast, our study uniquely addresses adversarial robustness in structured, tabular data, introducing an adversarial in-context learning approach that improves robustness by modifying only context data, thus avoiding parameter updates.

%% file: doc/02_problem.tex
\section{Problem Formulation}

\subsection{\textcolor{black}{Threat Model}}\label{sec:threats}

We set ourselves to study the robustness of Tabular FMs in the classification settings. Without loss of generality, this is described by defining $X_{context} \in \mathcal{R}^{(n \times D)}$ some context tabular data and their corresponding labels $Y_{context} \in \{1, ..., C\}^{n}$ ($C$ number of classes, $n$ number context data, $D$ number of tabular features) and $X_{infer} \in \mathcal{R}^{(m \times D)}$ ($m$ number of data to infer on) some tabular data to infer on. We define a Tabular FM as $h(. | X_{context}, Y_{context}, \theta)$ and the classification task aims to infer the labels of $X_{infer}$.

In that setting, we assume an adversary whose goal is to attack our Tabular FM through an evasion attack, in order to modify the inferred label of any $X_{infer}$ to a different one. Formally, given an input $x \in X_{infer}$, the objective is to generate $x + \delta$, where $\delta$ is a perturbation, such as $h(x | X_{context}, Y_{context}, \theta) \neq h(x + \delta | X_{context}, Y_{context}, \theta)$, $\delta < \epsilon$ where $\epsilon$ is the maximal perturbation allowed (e.g. so the attack is imperceptible) and $x + \delta$ follows the set of constraints $\Omega$ of the task. For instance, in a binary loan approval task, the adversary might want to modify any tabular input to flip the label to $1$ (approved), while keeping the features constrained (e.g. salary can not be negative etc.). 

We assume the worst settings in our case, that is, the adversary has white-box access to the model and so can access the weights AND the context of the Tabular FM, either directly or indirectly (e.g., through an exposed API for models deployed online). As we do not tackle backdoor or poisoning attacks, we suppose that neither the Tabular FM under attack itself, nor its context, is corrupted at any point, only the $X_{infer}$ are affected. The goal of the defender is to prevent the label from being flipped. We focus on potential hardening techniques via re-training.

In all cases, we used an $\epsilon = 0.5$ as the attack distance budget and $\epsilon = 0.3$ as the defence distance budget. This follows a classical threat scenario where we suppose that the adversary has a similar or higher perturbation budget than the defender \cite{simonetto2024tabularbench, madry2017towards}.

\subsection{\textcolor{black}{Tabular Adversarial Attacks}}\label{sec:tab-attack}

In our study, we focus on constrained adversarial attacks following the definition of~\cite{simonetto2022unified} to generate adversarial examples that respect domain constraints.
Following recommendations by Croce et al. \cite{croce2020robustbench} and Carlini et al. \cite{carlini2019evaluating}, our selection of attacks covers gradient-based and search-based attacks. 

The first attack we use is Constrained Adaptive Projected Gradient Descent (CAPGD)~\cite{simonetto2024constrained}, which is an adaptation of PGD on constrained (tabular) data. Given an input $x$ and its label $y$, PGD iteratively modify the input with a perturbation $\delta$ following the sign of the gradient $\nabla_{x}$, that is at iteration $t + 1$:

\begin{equation}
    x^{t+1} = \Pi_{x+\delta} (x^{t} + \eta \times sgn(\nabla_{x} \mathcal{L}(\theta_h, x^{t}, y))),
\end{equation}

where $\Pi_{x+\delta}$ is a clip function ensuring the that $x + \delta$ stays within bound,  $\alpha$ is a step parameter and $\nabla_x\mathcal{L}$ is the gradient of the loss for our task.

As PGD formulation does not handle constraints, it was extended via CAPGD to acknowledge them. Formally, CAPGD translates each constraint $\omega_i \in \Omega$, expressed in a constraint language, of the data $x$ into a differentiable function $penalty(x, \omega_i)$ using a translation table \cite{simonetto2022unified}. This results in a new loss objective 
\begin{equation*}
    \mathcal{L}^*(\theta_h, x^{t}, y, \Omega) - \sum_{\omega_i \in \Omega} penalty(x, \omega_i).
\end{equation*}

On top of that, it further introduces: 1) A repair operator $R_\Omega$ that projects back the example produced at each iteration into the constraint-abiding data space, 2) an adaptive step size based on the loss at defined checkpoints, so $\eta = \eta^{t}(\epsilon, \mathcal{L})$. All in all, the iterative scheme of CAPGD can be written as:

\begin{equation}
    x^{t+1} = R_\Omega(\Pi_{x+\delta} (x^{t} + \eta^{t} \times sgn(\nabla_{x} \mathcal{L}^*(\theta_h, x^{t}, y, \Omega)))).
\end{equation}

We refer the reader to the original paper~\cite{simonetto2024constrained} for more information on the attack.

The second attack we use is Constrained Adaptive Attack (CAA)~\cite{simonetto2024constrained}, which is a combination of both the previously introduced CAPGD and MOEVA~\cite{simonetto2022unified}. CAA successively apply CAPGD first and then MOEVA, benefiting first from CAPGD's fast, valid input generation before resorting to a slower but more effective MOEVA search-based approach. MOEVA is a multiobjective search attack adapted for tabular data and constitutes the state-of-the-art for search-based evasion attacks on tabular data when considering domain constraints. MOEVA is based on a multi-objective algorithm (R-NSGA-III) with three objectives to optimise. The first objective aims to cause misclassification by minimising the prediction probability $h(x|X_{context}, Y_{context}, \theta)$. The second objective aims to minimise the perturbation between $\delta$ between original and adversarial examples. Finally, the last objective is to satisfy domain constraints using the penalty function $\sum_{\omega_i \in \Omega} penalty(x, \omega_i)$. These three objectives form the fitness function used in the genetic algorithm.

Finally, as a point of comparison, we use as a third attack the identity attack, i.e. using the clean data as is. This serves as a control in our experiments to determine whether our studied mechanisms have a concrete impact and are not just an artefact of the data.

\subsection{\textcolor{black}{Metrics}}\label{sec:metrics}

Given the adversarial setting and that all our datasets are binary, we consider the positive class of interest as done in previous studies~\cite{simonetto2024constrained} and so the aim of the adversary is to flip the label of any input $x$ to be positive. As such, we will track the robust accuracy, that is, the accuracy with regard to the positive class in all our experiments unless specified otherwise. This is equivalent to computing the Recall on a binary task. Given $N$ true positive class of adversarial inputs with ground truth $y_{1, true}$ and their corresponding predictions through the model $y_{pred}$, we define the adversarial accuracy in our experiments as follows: $\frac{1}{N}\sum_i [y_{{1, true}_i} = y_{{pred}_i}]$.

%% file: doc/03_experimental_protocol.tex
\section{Experimental Setup}\label{sec:exp_setup}

\paragraph{Models}

In our experiments, we used the two current state-of-the-art tabular FMs, i.e. TabPFN (version 2) ~\cite{hollmann2025tabpfn}, and TabICL~\cite{qu2025tabicl}. To contrast our results, we used seven non-foundational models used in the literature, namely: TabTransformer~\cite{huang2020tabtransformer}, RLN~\cite{shavitt2018regularization}, VIME~\cite{yoon2020vime}, STG~\cite{icml2020_5085}, TabNet~\cite{arik2021tabnet}, XGBoost~\cite{XGBOOST} and RF~\cite{choi2011tree}. The FM are used with their default parameters unless stated otherwise. We use 1 estimator in the pre-processing step of both TabPFNv2 and TabICL (instead of 4). While this is not the exact scheme used in TabPFNv2 and TabICL original implementation, this allows us to have better control over the attack/defence of the model. In practice, we did not find any significant difference between using 1 or the default 4 estimators for clean accuracy. For non-FM, we reuse the parameters and models of TabularBench~\cite{simonetto2024tabularbench} that were trained with 100 iterations of hyperparameter search.

\paragraph{Datasets}

Regarding datasets,\footnote{The use of these datasets in this study is intended for scientific evaluation only, without implying support for their practical applications} we chose three tabular datasets introduced in TabularBench~\cite{simonetto2024tabularbench} benchmark. The first one is WIDS~\cite{wids}, a medical dataset on the survival of patients admitted to the ICU, with 186 features and 31 linear domain constraints. The second is LCLD~\cite{lcld}, a credit-scoring dataset containing accepted and rejected credit requests with 28 features and
9 non-linear constraints. URL~\cite{hannousse2021towards} is a dataset comprising both legitimate and phishing URLs with 63 features and 14 domain constraints. These three datasets were chosen as they have been extensively used in the literature and are also applicable to both tabular foundational models under study. In fact, TabPFNv2 and TabICL feature context can handle only up to 500 features, which disqualified other TabularBench datasets, such as Malware~\cite{dyrmishi2023empirical} (24,222 features) or CTU~\cite{chernikova2022fence} (756 features).

\paragraph{Model Training}

Tabular FM in our experiments only accept a context of up to 10k, as per model limitations. Although TabICL theoretically can go to 500k, the model does not scale beyond 10k in our experiments. As such, since our used datasets have a size > 10k, a subsampling was necessary. To avoid poor sampling, we sample with 10 seeds and, with and without rebalancing, fitted a TabPFNv2 model that we evaluated on the validation set of each dataset with the Matthew Correlation Coefficient (MCC). We kept the best seed and sampling method throughout all further training / fine-tuning. For non-FM, we used the training protocol and models used in TabularBench \cite{simonetto2024tabularbench}.

\paragraph{Adversarial Attacks}

We used the attacks described in Section~\ref{sec:tab-attack}. In all cases, we left the default parameters as used in the original attack, i.e. $10$ iterations for CAPGD and $100$ iterations for MOEVA.

%% file: doc/04_rq1.tex
\input{tables/rq1_robust_accuracy}

\section{Robustness of Tabular FM to Evasion Attacks}

\subsection{Experimental Setup}

Our first investigation explores the out-of-the-box robust performance of recent tabular FM compared to common tabular non-FM models. To do so, we used the models and datasets described in Section~\ref{sec:exp_setup}. For each of the combinations, we apply the identity, CAPGD and CAA attacks on the positive label test data of each respective dataset as described in Section~\ref{sec:tab-attack} and Section~\ref{sec:metrics}. For CAPGD and CAA, we do so over $5$ seeded runs to account for nondeterminism in the attack generation methods. We summarise in Table~\ref{tab:rq1_results} the adversarial accuracy (recall) against clean and adversarial attacks.

\subsection{Experimental Results}\label{sec:rq1-experimental-protocol}

Focusing first on clean data, \textcolor{black}{TabFMs generally overperform non-FMs when it comes to MCC. However, when considering recall,} our results demonstrate that TabPFNv2, and TabICL consistently underperform on LCLD and WIDS datasets, both traditional deep tabular models and tree-based XGB models. They only outperform deep tabular models on the simple URL dataset. \textcolor{black}{We attribute this discrepancy due to our datasets: LCLD and WIDS are both unbalanced, while URL is balanced which might explain why TabFMs (which were not trained on the data) struggle when it come to the positive class.} TabTransformer is the best model for the LCLD (70.2\%) dataset, XGBoost remains the best for WIDS (80.4\%) and URL (97.4\%).

Moving to adversarial robustness, most models are poorly robust, with WIDS being the dataset where models tend to be the most robust on average. STG is the tabular non-FMs model that is the most robust across the board, while TabNet is the model with the lowest robustness. Tabular FMs models are generally not robust.
% with TabICL having the best performance. 

\textbf{Conclusion:} \textit{All in all, our evidence shows that tabular models are vulnerable to evasion attacks. Tabular FMs models are generally more vulnerable than the best-performing tabular non-FMs models.}

We extend the study by examining the impact of hyperparameters ($\epsilon$ of CAPGD, iteration budget of CAPGD/MOEVA and context size) on the robustness obtained for tabular FMs. \textcolor{black}{We give in Appendix \ref{sec:app-exp-budget-classical} the results obtained for non-FMs.}

\paragraph{Impact of perturbation distance $\epsilon$}
\label{sec:epsilon-nonrobust}

\begin{figure*}[ht]
    \centering
    \begin{subfigure}[t]{0.49\textwidth}
        \centering
        \includegraphics[width=0.8\linewidth]{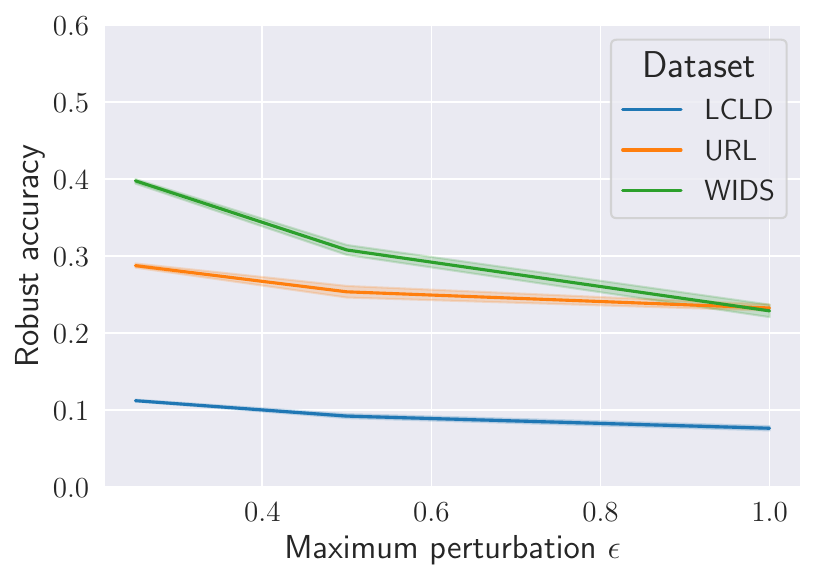}
        \caption{TabPFNv2}
        \label{fig:eps_tabpfnv2}
    \end{subfigure}%
    \hfill
    \begin{subfigure}[t]{0.49\textwidth}
        \centering
        \includegraphics[width=0.8\linewidth]{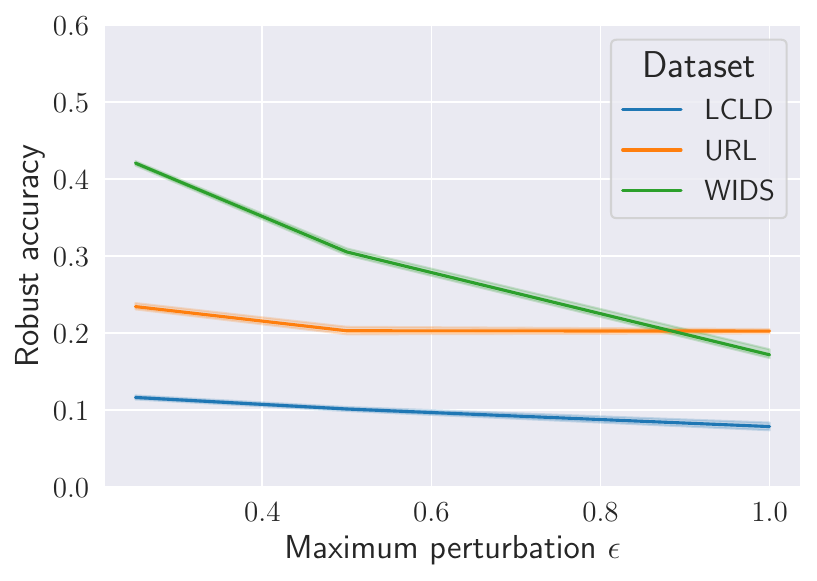}
        \caption{TabICL}
        \label{fig:eps_tabicl}
    \end{subfigure}
    \caption{Robust accuracy with CAA with varying maximum perturbation $\epsilon$.}
    \label{fig:eps}
\end{figure*}

We modify the value of the strength of the perturbation distance $\epsilon$ of the CAA attack as used in RQ1. The values studied are \{0.25, 0.5, 1.0\} allowing us to examine lower/equal/higher perturbation budget for the adversary and following previous works~\cite{simonetto2024constrained}. We give the adversarial accuracy for both TabPFNv2 and TabICL when using CAA on Figure~\ref{fig:eps}. We only experimented with CAA as CAPGD is a step within CAA, and thus, the results obtained for CAA will be representative of the results we would obtain on CAPGD.

As can be observed in the Figure, for both models, $\epsilon$ has a low to no impact on the robust accuracy obtained (except on WIDS). If initially there is some decrease between $\epsilon = 0.25$ and $\epsilon = 0.5$, the robust accuracy flattens when going between $\epsilon = 0.5$ and $\epsilon = 1.0$. Only in the case of the WIDS dataset does the robust accuracy keep on decreasing, albeit at a slower rate.

\paragraph{Impact of perturbation budget}
\label{sec:budget-nonrobust}

\begin{figure*}[ht]
    \centering
    \begin{subfigure}[t]{0.49\textwidth}
        \centering
        \includegraphics[width=0.8\linewidth]{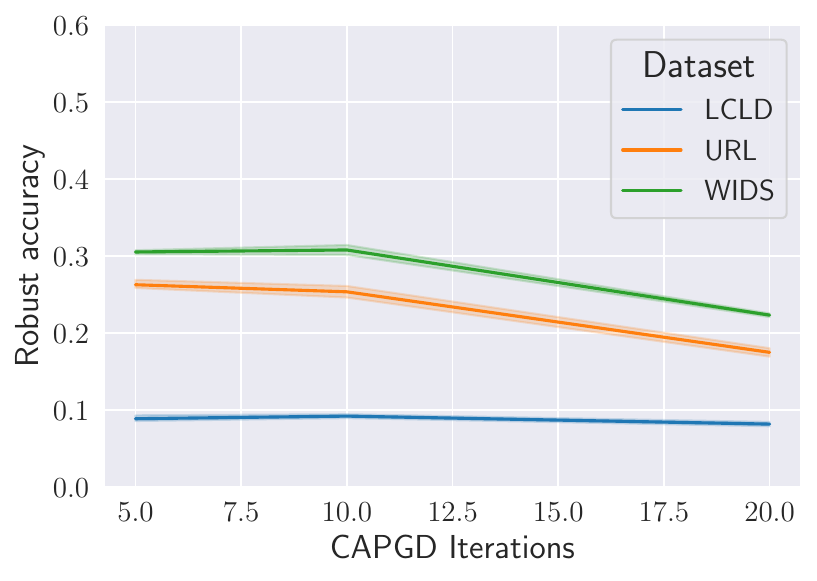}
        \caption{TabPFNv2}
        \label{fig:n_steps_tabpfnv2}
    \end{subfigure}%
    \hfill
    \begin{subfigure}[t]{0.49\textwidth}
        \centering
        \includegraphics[width=0.8\linewidth]{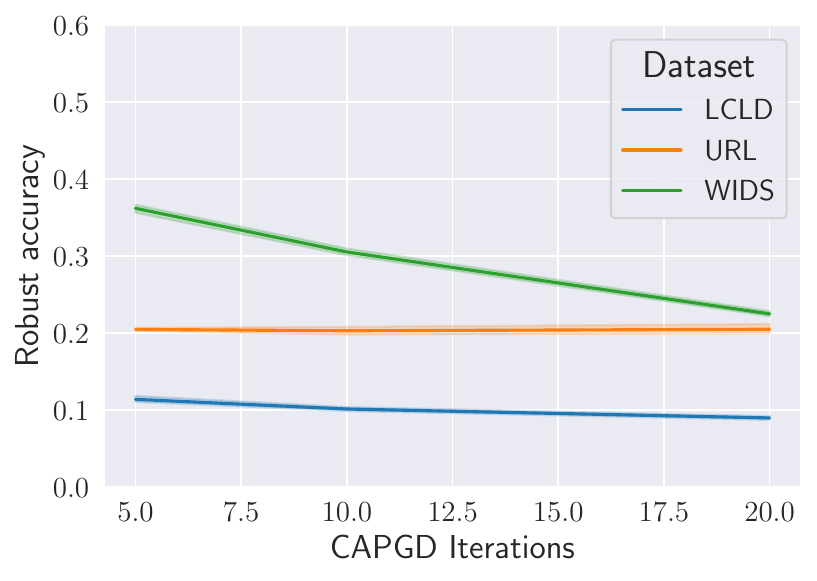}
        \caption{TabICL}
        \label{fig:n_steps_tabicl}
    \end{subfigure}
    \caption{Robust accuracy with CAA with varying search attack iterations in CAPGD.}
    \label{fig:n_steps}
\end{figure*}

\begin{figure*}[ht]
    \centering
    \begin{subfigure}[t]{0.49\textwidth}
        \centering
        \includegraphics[width=0.8\linewidth]{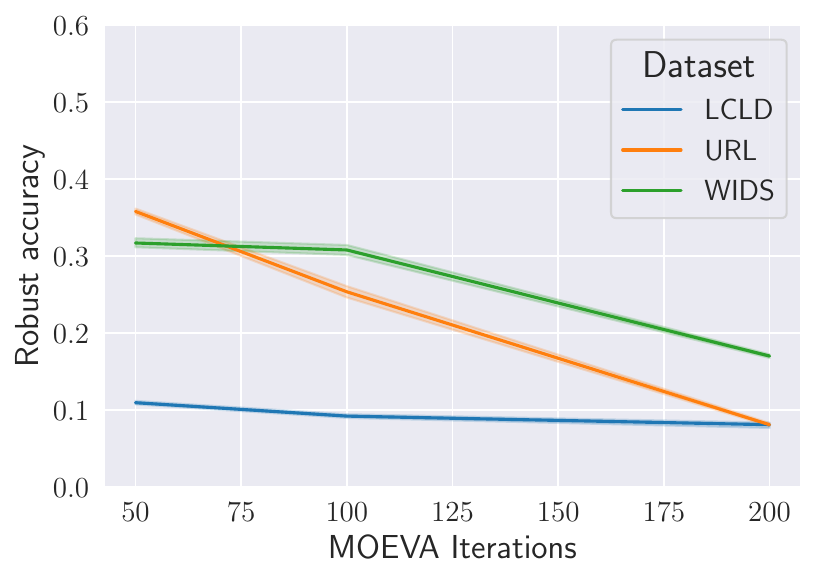}
        \caption{TabPFNv2}
        \label{fig:n_gen_tabpfnv2}
    \end{subfigure}%
    \hfill
    \begin{subfigure}[t]{0.49\textwidth}
        \centering
        \includegraphics[width=0.8\linewidth]{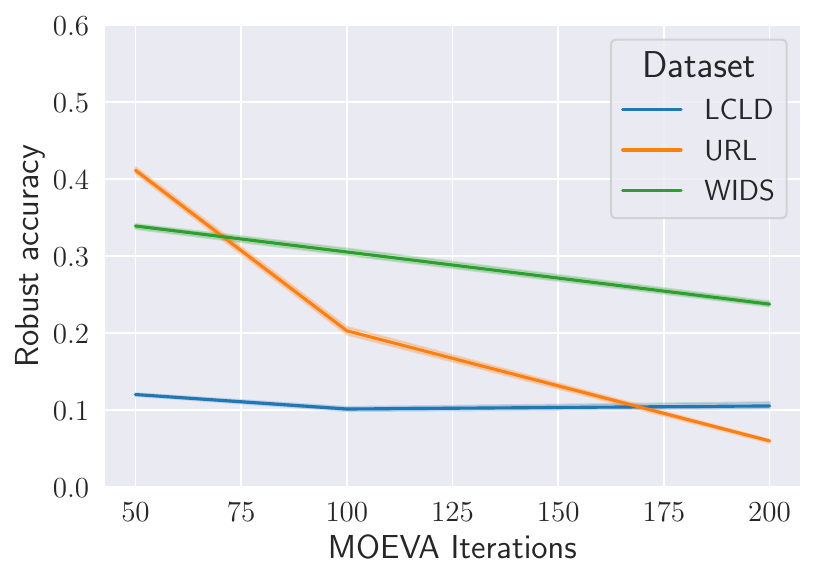}
        \caption{TabICL }
        \label{fig:n_gen_tabicl}
    \end{subfigure}
    \caption{Robust accuracy with CAA with varying search attack iterations in MOEVA.}
    \label{fig:n_gen}
\end{figure*}

We then study the impact of the perturbation budget on two axes: 1) The number of iterations within the CAPGD attack, 2) The number of iterations within the MOEVA search. Similarly to the previous experiment, we only study CAA as it is a composition of both attacks. For the number of iterations of CAPGD, we used \{5, 10, 20\}, and for the search iterations of MOEVA, we used \{50, 100, 200\}, in order to study a lower/equal/higher budget, following previous studies~\cite{simonetto2024constrained}. Robust accuracy obtained for both TabPFNv2 and TabICL can be found in Figure~\ref{fig:n_steps} for CAPGD iterations and in Figure~\ref{fig:n_gen} for MOEVA iterations.

Regarding the number of iterations within CAPGD, we see that it has little impact on the robust accuracy obtained, with 20 iterations providing marginally better results, besides on WIDS / TabICL. As such, the number of iterations on CAPGD does not seem to impact much the robust accuracy for both models. On the contrary, when analysing the number of iterations in the MOEVA search, we observe a decrease for both models on both URL and WIDS datasets, with the robust accuracy more than halving in both cases between $50$ and $200$ iterations for URL. However, we observe no meaningful variation when it comes to LCLD.

\paragraph{Impact of context size}
\label{sec:context-nonrobust}

\begin{figure*}[ht]
    \centering
    \begin{subfigure}[t]{0.49\textwidth}
        \centering
        \includegraphics[width=0.8\linewidth]{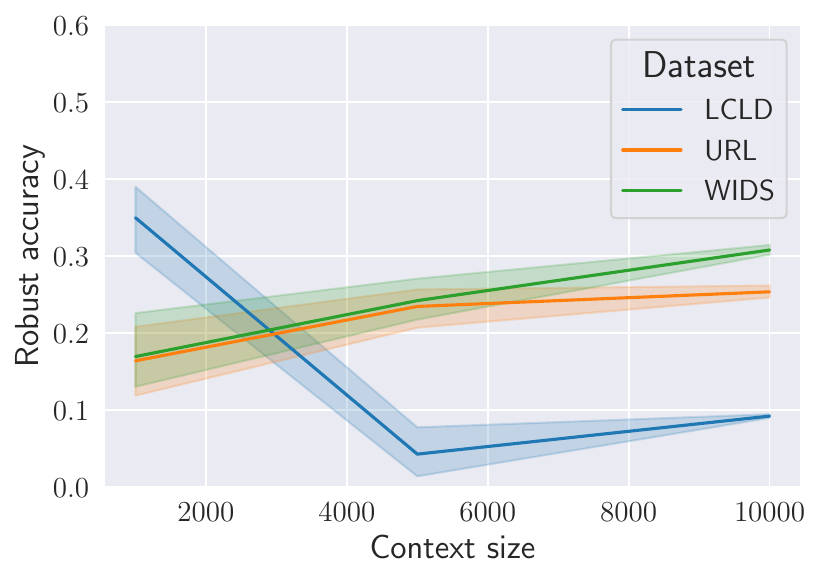}
        \caption{TabPFNv2}
        \label{fig:subset_tabpfnv2}
    \end{subfigure}%
    \hfill
    \begin{subfigure}[t]{0.49\textwidth}
        \centering
        \includegraphics[width=0.8\linewidth]{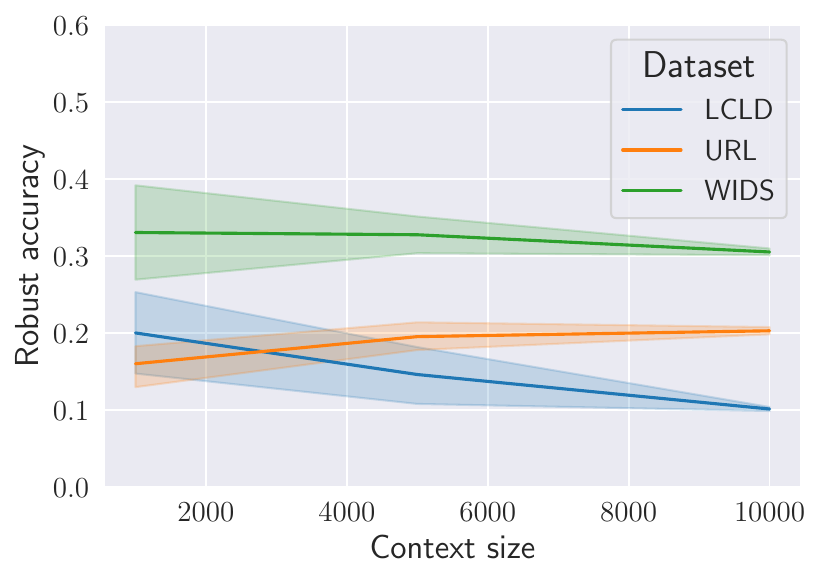}
        \caption{TabICL}
        \label{fig:subset_tabicl}
    \end{subfigure}
    \caption{Robust accuracy with CAA with varying model context size.}
    \label{fig:subset}
\end{figure*}

We finally analysed the impact of the context size on the robust accuracy. To do so, we modify the context size by taking as values \{1000, 5000, 10000\}, to have a lower/equal/higher context size. Same as before, we only look at the results for the strongest attack CAA. Results are given in Figure~\ref{fig:subset}.

We note that, generally, modifying the context size has little effect on both models. We do note a slight tendency to increase the robust accuracy when increasing the context size for TabPFNv2 on URL and WIDS, and for TabICL on URL. The sharp difference we observe between the models mainly comes from the LCLD dataset, where robust accuracy does not change much for TabICL, there is a sharp decline in the robust accuracy of TabPFNv2 when increasing the context. \textcolor{black}{We attribute this phenomenon to LCLD size (over a million examples), and so the sampled context might cause some overfitting in the low sample regime.}

%% file: tables/rq1_robust_accuracy.tex
\begin{table*}[t]
\centering
\caption{Robust accuracy against Clean data, CAPGD and CAA adversarial examples. \textcolor{black}{For Clean data we also give Mathew Correlation Coefficient (MCC) in parenthesis}. A lower robust accuracy means a more effective attack. CAPGD is not applicable to XGB and RF as it is pure gradient attack.}
\label{tab:rq1_results}
\small

\resizebox{0.8\textwidth}{!}{\begin{tabular}{lrrrrrrrrr}
\toprule
Dataset & \multicolumn{3}{c}{LCLD} & \multicolumn{3}{c}{URL} & \multicolumn{3}{c}{WIDS} \\ %& \multicolumn{3}{c}{HELOC}\\

\cmidrule(lr){2-4} \cmidrule(lr){5-7} \cmidrule(lr){8-10}% \cmidrule(lr){11-13}

Model & Clean & CAPGD & CAA & Clean & CAPGD & CAA & Clean & CAPGD & CAA \\ %& Clean & CAPGD & CAA \\
\midrule
% \midrule
RLN & 68.1 (25.3) & 0.2\tiny{$\pm 0.1$} & 0.0\tiny{$\pm 0.1$} & 94.4 (89.1) & 11.5\tiny{$\pm 0.3$} & 10.6\tiny{$\pm 0.4$} & 77.5 (37.7) & 62.1\tiny{$\pm 0.4$} & 60.9\tiny{$\pm 0.4$} \\ %& 69.4 & 3.1 \tiny{$\pm 0.1$} & 0.0 \tiny{$\pm 0.0$}\\
STG & 66.0 (24.4) & 56.1\tiny{$\pm 0.3$} & 52.6\tiny{$\pm 0.3$} & 93.3 (83.9) & 71.5\tiny{$\pm 0.2$} & 56.9\tiny{$\pm 0.3$} & 77.6 (36.3) & 64.2\tiny{$\pm 0.6$} & 63.6\tiny{$\pm 0.7$} \\ %& 43.4 & 1.4 \tiny{$\pm 0.2$} & 0.0 \tiny{$\pm 0.0$}\\
TabNet & 67.5 (26.0) & 7.3\tiny{$\pm 0.4$} & 0.3\tiny{$\pm 0.2$} & 93.4 (89.2) & 17.9\tiny{$\pm 0.9$} & 9.1\tiny{$\pm 0.4$} & 79.7 (36.7) & 7.9\tiny{$\pm 0.3$} & 5.5\tiny{$\pm 0.5$} \\ %& 72.9 & 8.1 \tiny{$\pm 0.3$} & 6.1 \tiny{$\pm 0.4$}\\
TabTr. & 70.2 (25.2) & 36.9\tiny{$\pm 0.9$} & 5.6\tiny{$\pm 0.4$} & 93.6 (88.0) & 9.4\tiny{$\pm 0.2$} & 8.3\tiny{$\pm 0.2$} & 75.5 (38.5) & 47.5\tiny{$\pm 0.6$} & 46.0\tiny{$\pm 0.5$} \\ %& 75.5 & 0.0 \tiny{$\pm 0.0$} & 0.0 \tiny{$\pm 0.0$}\\
VIME & 67.1 (24.9) & 3.6\tiny{$\pm 0.4$} & 2.8\tiny{$\pm 0.4$} & 92.5 (85.6) & 54.4\tiny{$\pm 0.2$} & 47.1\tiny{$\pm 1.3$} & 72.3 (38.7) & 50.7\tiny{$\pm 0.3$} & 50.3\tiny{$\pm 0.3$} \\ %& 71.4 & 9.6 \tiny{$\pm 0.1$} & 9.4 \tiny{$\pm 0.1$}\\
RF & 64.6 (25.6) & N/A\tiny & 27.5\tiny{$\pm 0.6$} & 95.9 (90.3) & N/A\tiny & 48.3\tiny{$\pm 1.6$} & 49.3 (43.8) & N/A\tiny & 11.0\tiny{$\pm 0.4$} \\ %& 72.5 & N/A & 6.9 \tiny{$\pm 0.3$}\\
XGB & 67.9 (26.1) & N/A\tiny & 10.3\tiny{$\pm 0.7$} & 97.4 (93.8) & N/A\tiny & 16.3\tiny{$\pm 0.9$} & 80.4 (40.5) & N/A\tiny & 35.7\tiny{$\pm 0.7$} \\ %& 69.6 & N/A & 2.1 \tiny{$\pm 0.2$}\\
\rule{1pt}{0ex} \\

% TabPFNv2 & 59.6 (25.4) & 17.3 \tiny{$\pm 0.1$} & 8.5 \tiny{$\pm 0.0$} & 97.0 (92.8) & 42.4 \tiny{$\pm 0.2$} & 23.7 \tiny{$\pm 0.3$} & 70.5 (43.0) & 33.1 \tiny{$\pm 0.3$} & 26.6 \tiny{$\pm 0.3$} 
% %& 73.4 & 0.5 \tiny{$\pm 0.1$} & 0.0 \tiny{$\pm 0.0$} 
% \\
TabPFNv2 & 59.7 (25.4) & 22.4 \tiny{$\pm 0.1$} & 9.0 \tiny{$\pm 0.2$} & 97.1 (92.8) & 47.6 \tiny{$\pm 0.2$} & 25.2 \tiny{$\pm 0.5$} & 70.7 (43.0) & 49.2 \tiny{$\pm 0.4$} & 30.8 \tiny{$\pm 0.4$} \\
% & 73.5 & 5.0 \tiny{$\pm 0.6$} & 0.3 \tiny{$\pm 0.1$} \\

% TabPFNv2 (Precision)& 59.7 & 22.4 \tiny{$\pm 0.1$} & 9.1 \tiny{$\pm 0.1$} & 97.1 & 47.6 \tiny{$\pm 0.2$} & 25.2 \tiny{$\pm 0.5$} & 70.7 & 49.2 \tiny{$\pm 0.4$} & 30.8 \tiny{$\pm 0.4$} \\
%& 73.5 & 5.0 \tiny{$\pm 0.6$} & 0.3 \tiny{$\pm 0.1$} \\

\rule{1pt}{0ex} \\

% TabICL (25.5)& 59.6 & 18.3 \tiny{$\pm 0.3$} & 9.9 \tiny{$\pm 0.2$} & 96.7 & 82.8 \tiny{$\pm 0.2$} & 20.4 \tiny{$\pm 0.3$} & 75.5 & 38.4 \tiny{$\pm 0.1$} & 29.9 \tiny{$\pm 0.1$}  \\
% & 70.9 & 11.9 \tiny{$\pm 0.2$} & 10.9 \tiny{$\pm 0.1$} 

% TabICL & 59.6 & 18.5 \tiny{$\pm 0.4$} & 9.9 \tiny{$\pm 0.1$} & 96.7 & 82.9 \tiny{$\pm 0.2$} & 20.2 \tiny{$\pm 0.3$} & 75.5 & 38.2 \tiny{$\pm 0.3$} & 30.5 \tiny{$\pm 0.3$} & 70.9 & 11.9 \tiny{$\pm 0.1$} & 11.0 \tiny{$\pm 0.1$} \\

TabICL& 59.6 (25.5) & 18.5 \tiny{$\pm 0.4$} & 10.2 \tiny{$\pm 0.1$} & 96.7 (92.8) & 82.9 \tiny{$\pm 0.2$} & 20.2 \tiny{$\pm 0.3$} & 75.5 (42.0) & 38.2 \tiny{$\pm 0.3$} & 30.5 \tiny{$\pm 0.2$} \\ %& 70.9 & 11.9 \tiny{$\pm 0.1$} & 11.0 \tiny{$\pm 0.1$} \\

\bottomrule
\end{tabular}}
\end{table*}

%% file: doc/05_rq2.tex
\input{tables/rq3.1_transferability_inter_arch}

\section{Transfer Attacks on and with Tabular FM}\label{sec:rq2}

\subsection{Experimental Setup}

In adversarial machine learning, black-box attacks often rely on transferability, the phenomenon in which adversarial examples crafted on a surrogate model are applied against a different, unseen
target model. Given the versatility of tabular FM and their generalization capabilities, we next explore the benefits of using these FM as a surrogate to generate transferable attacks. Transferability can be studied given two axes:

\begin{itemize}
    \item \textit{Inter-family transferability}: Adversarial examples crafted on tabular FM are applied to traditional models (both tree-based and deep-learning-based), and vice-versa.
    \item \textit{Intra-family transferability}: Adversarial examples crafted on one type of model (FM or classical) are applied to the same type.
\end{itemize}

\subsection{Experimental Results}\label{sec:rq2-experimental-results}

% \textbf{Tabular Foundation Models are not a reliable surrogate for transferable attacks:} 
We reuse the adversarial examples obtained in the previous experiments using CAA for the $5$ seeds and apply them to other models. We then average the robust accuracy obtained across the seeds. 
We aggregate in Table~\ref{tab:transferability_inter_arch} the robust accuracy under different combinations of source and target tabular models.

Overall, most classical models remain good surrogates to attack FM, and tabular FM are not necessarily more robust against inter-family transfer than intra-family transfer. For instance, we can highlight that, on LCLD, attacking TabPFNv2 or TabICL with a transfer attack using VIME is as effective as running the CAA attack on these FM. On the other hand, tabular FM can be used as a surrogate to attack other classical models. In particular, tree-based models such as RF and XGB can be more vulnerable to attack from FM models than other classical models or even other tree models. For example, on WIDS, adversarial attacks crafted on TabICL achieve 27.7\% on RBF and 56.7\% on XGB, among the best surrogates, even better than other tree models. Transferability is weakly susceptible to the seed of the attack, yet it is much more susceptible to the dataset and the architecture of the source/target models. Yet this is in line with general research on transferability (between non-foundation models in computer vision, see e.g. \cite{waseda2023closer}). 

\textbf{Conclusion:} \textit{Tabular FM are vulnerable to transfer attacks and can be effective surrogates to attack classical models, though we observe large variance in the transfer attack success rate.}

\input{tables/rq3.2_transferability_foundation}

% Tabular FM do not bring added robustness against transferable attacks and so are not a reliable surrogate for transferable attacks.

% \begin{figure*}[t]
%     \centering
%     \begin{subfigure}[t]{0.49\textwidth}
%         \centering
%         \includegraphics[width=\linewidth]{fig/source_tabpfn_target_architecture.pdf}
%         \caption{Difference between the CAA attack and TabPFNv2 as a source in transferability settings across all datasets and seeds.}
%         \label{fig:source_tabpfn_dataset}
%     \end{subfigure}%
%     \hfill
%     \begin{subfigure}[t]{0.49\textwidth}
%         \centering
%         \includegraphics[width=\linewidth]{fig/source_tabpfn_dataset.pdf}
%         \caption{Difference between the CAA attack and TabPFNv2 as a source in transferability settings across all target architectures and seeds.}
%         \label{fig:source_tabpfn_target_architecture}
%     \end{subfigure}
%     \caption{Comparison of TabPFNv2 as a source in transferability settings.}
%     \label{fig:tabpfn_comparison}

% \end{figure*}

Next, we evaluate the sensitivity of the transferable attacks for the same tabular FM as the source and target. \textcolor{black}{The goal is to represent an adversary using transferability to attack a model, but in a more realistic scenario, i.e. with only \textit{some} knowledge of the data/target. We simulate this via two scenarios.} In the first scenario, we subsample the context of the source by taking 10\% of the original context of the dataset before calculating the robust accuracy. In the second scenario, we build a context of 10k examples from the same original full training set. This is only possible for LCLD and WIDS datasets, whose original training sets' size exceeds the context used in the target model. We compare these scenarios with the case where the surrogate and target share the same context and distribution (\textcolor{black}{i.e. adversary has full knowledge)}. Results are given in Table \ref{tab:transferability_foundation}. On WIDS, there is a marginal benefit of collecting a full, exact context compared to only a subset. In addition, having only access to the distribution yields similar performances to the exact context for WIDS. On LCLD, collecting more examples of the same distribution is more effective than obtaining exactly the target's context. On URL, having the exact context is more beneficial than having only a subset. 

\textbf{Conclusion:} \textit{Context and distribution have an impact on the transferability of attacks for tabular FM: \textcolor{black}{limited knowledge of the context might be enough to attack a tabular FM.}}

% \begin{figure}
%     \centering
%     \includegraphics[width=1\linewidth]{fig/source_tabpfn_target_architecture.pdf}
%     \caption{Difference between white-box attack and TabPFNv22 as a source in transferability settings accross all datasets and seeds.}
%     \label{fig:enter-label}
% \end{figure}

% \begin{figure}
%     \centering
%     \includegraphics[width=1\linewidth]{fig/source_tabpfn_dataset.pdf}
%     \caption{Difference between white-box attack and TabPFNv22 as a source in transferability settings accross all target architecture and seeds.}
%     \label{fig:enter-label}
% \end{figure}

%% file: tables/rq3.1_transferability_inter_arch.tex
\begin{table*}[t]
\centering
\caption{Transferability of evasion attacks (CAA) across architectures. We report the mean and standard deviation of the robust accuracy across seeds. }\label{tab:transferability_inter_arch}
\small

\begin{tabular}{llrrrrrrrrr}
\toprule
 & & \multicolumn{9}{c}{Target}\\
 \rule{0pt}{3ex} 
Dataset & Source &  RLN & STG & TabNet & TabTr. & VIME & RF & XGB & TabPFNv2 & TabICL \\

\midrule
\multirow[c]{10}{*}{LCLD} & Clean & 68.1 & 66.0 & 67.5 & 70.2 & 67.1 & 64.6 & 67.9 & 59.6 & 59.6 \\
 & RLN & 0.0\tiny{$\pm 0.1$} & 63.7\tiny{$\pm 0.0$} & 5.5\tiny{$\pm 0.1$} & 14.1\tiny{$\pm 0.2$} & 7.9\tiny{$\pm 0.2$} & 7.8\tiny{$\pm 0.1$} & 11.8\tiny{$\pm 0.1$} & 19.9\tiny{$\pm 0.1$} & 22.6\tiny{$\pm 0.2$} \\
 & STG & 58.0\tiny{$\pm 0.2$} & 52.6\tiny{$\pm 0.3$} & 58.9\tiny{$\pm 0.4$} & 59.4\tiny{$\pm 0.3$} & 56.4\tiny{$\pm 0.3$} & 55.4\tiny{$\pm 0.1$} & 58.5\tiny{$\pm 0.2$} & 52.1\tiny{$\pm 0.3$} & 52.2\tiny{$\pm 0.3$} \\
 & TabNet & 12.3\tiny{$\pm 0.3$} & 65.1\tiny{$\pm 0.1$} & 0.3\tiny{$\pm 0.2$} & 52.1\tiny{$\pm 0.2$} & 47.2\tiny{$\pm 0.4$} & 28.1\tiny{$\pm 0.3$} & 24.2\tiny{$\pm 0.5$} & 30.1\tiny{$\pm 0.2$} & 32.7\tiny{$\pm 0.4$} \\
 & TabTr. & 8.4\tiny{$\pm 0.5$} & 59.8\tiny{$\pm 0.2$} & 13.8\tiny{$\pm 0.5$} & 5.6\tiny{$\pm 0.4$} & 20.5\tiny{$\pm 0.8$} & 21.3\tiny{$\pm 0.3$} & 29.1\tiny{$\pm 0.6$} & 11.7\tiny{$\pm 0.6$} & 13.9\tiny{$\pm 0.5$} \\
 & VIME & 7.6\tiny{$\pm 0.4$} & 62.1\tiny{$\pm 0.1$} & 9.3\tiny{$\pm 0.5$} & 14.1\tiny{$\pm 0.5$} & 2.8\tiny{$\pm 0.4$} & 6.5\tiny{$\pm 0.2$} & 10.2\tiny{$\pm 0.3$} & 9.9\tiny{$\pm 0.1$} & 8.8\tiny{$\pm 0.3$} \\
 & RF & 40.3\tiny{$\pm 0.2$} & 63.8\tiny{$\pm 0.2$} & 45.8\tiny{$\pm 0.9$} & 53.1\tiny{$\pm 0.7$} & 53.4\tiny{$\pm 0.8$} & 27.5\tiny{$\pm 0.6$} & 42.1\tiny{$\pm 0.5$} & 34.6\tiny{$\pm 0.3$} & 35.8\tiny{$\pm 0.8$} \\
 & XGB & 30.3\tiny{$\pm 1.0$} & 63.8\tiny{$\pm 0.4$} & 37.0\tiny{$\pm 0.5$} & 51.1\tiny{$\pm 0.7$} & 50.7\tiny{$\pm 1.0$} & 39.2\tiny{$\pm 0.6$} & 10.3\tiny{$\pm 0.7$} & 28.8\tiny{$\pm 0.5$} & 29.7\tiny{$\pm 0.7$} \\
     & TabPFNv2 & 24.7\tiny{$\pm 0.8$} & 65.0\tiny{$\pm 0.2$} & 30.4\tiny{$\pm 1.6$} & 50.2\tiny{$\pm 1.2$} & 47.5\tiny{$\pm 1.7$} & 30.8\tiny{$\pm 0.5$} & 32.7\tiny{$\pm 0.9$} & 9.2\tiny{$\pm 0.4$} & 22.7\tiny{$\pm 0.7$} \\
 & TabICL & 32.9\tiny{$\pm 0.5$} & 65.0\tiny{$\pm 0.4$} & 41.5\tiny{$\pm 1.4$} & 56.1\tiny{$\pm 0.5$} & 55.5\tiny{$\pm 0.7$} & 34.8\tiny{$\pm 1.2$} & 36.8\tiny{$\pm 0.6$} & 26.0\tiny{$\pm 1.3$} & 10.1\tiny{$\pm 0.4$} \\
\cline{1-11}
\multirow[c]{10}{*}{URL} & Clean & 94.4 & 93.3 & 93.4 & 93.6 & 92.5 & 95.9 & 97.4 & 97.0 & 96.7 \\
 & RLN & 10.6\tiny{$\pm 0.4$} & 84.3\tiny{$\pm 0.1$} & 37.5\tiny{$\pm 0.2$} & 13.9\tiny{$\pm 0.3$} & 86.6\tiny{$\pm 0.0$} & 76.3\tiny{$\pm 0.2$} & 57.0\tiny{$\pm 0.2$} & 63.0\tiny{$\pm 0.2$} & 17.5\tiny{$\pm 0.2$} \\
 & STG & 66.2\tiny{$\pm 0.1$} & 56.9\tiny{$\pm 0.3$} & 67.7\tiny{$\pm 0.5$} & 64.7\tiny{$\pm 0.2$} & 69.1\tiny{$\pm 0.4$} & 78.4\tiny{$\pm 0.5$} & 77.4\tiny{$\pm 0.7$} & 82.5\tiny{$\pm 0.8$} & 71.9\tiny{$\pm 1.0$} \\
 & TabNet & 64.0\tiny{$\pm 0.3$} & 87.7\tiny{$\pm 0.2$} & 9.1\tiny{$\pm 0.4$} & 59.9\tiny{$\pm 0.2$} & 92.3\tiny{$\pm 0.1$} & 87.5\tiny{$\pm 0.1$} & 77.9\tiny{$\pm 0.3$} & 68.3\tiny{$\pm 0.5$} & 72.9\tiny{$\pm 0.4$} \\
 & TabTr. & 17.1\tiny{$\pm 0.1$} & 86.5\tiny{$\pm 0.0$} & 40.5\tiny{$\pm 0.1$} & 8.3\tiny{$\pm 0.2$} & 86.1\tiny{$\pm 0.0$} & 79.9\tiny{$\pm 0.1$} & 46.6\tiny{$\pm 0.2$} & 69.3\tiny{$\pm 0.1$} & 18.1\tiny{$\pm 0.2$} \\
 & VIME & 59.5\tiny{$\pm 0.9$} & 81.3\tiny{$\pm 0.3$} & 72.9\tiny{$\pm 0.7$} & 54.1\tiny{$\pm 1.0$} & 47.1\tiny{$\pm 1.3$} & 80.5\tiny{$\pm 0.4$} & 72.9\tiny{$\pm 0.8$} & 90.0\tiny{$\pm 0.4$} & 69.1\tiny{$\pm 0.5$} \\
 & RF & 77.8\tiny{$\pm 0.9$} & 85.4\tiny{$\pm 0.5$} & 75.6\tiny{$\pm 1.1$} & 76.2\tiny{$\pm 0.5$} & 85.5\tiny{$\pm 0.6$} & 48.3\tiny{$\pm 1.6$} & 63.0\tiny{$\pm 1.2$} & 72.3\tiny{$\pm 0.7$} & 72.6\tiny{$\pm 2.0$} \\
 & XGB & 79.2\tiny{$\pm 0.3$} & 89.3\tiny{$\pm 0.5$} & 72.8\tiny{$\pm 1.2$} & 75.9\tiny{$\pm 0.5$} & 88.5\tiny{$\pm 0.3$} & 78.8\tiny{$\pm 1.1$} & 16.3\tiny{$\pm 0.9$} & 62.5\tiny{$\pm 0.9$} & 63.5\tiny{$\pm 0.7$} \\
 & TabPFNv2 & 81.3\tiny{$\pm 1.8$} & 90.0\tiny{$\pm 0.2$} & 70.5\tiny{$\pm 1.2$} & 66.4\tiny{$\pm 1.0$} & 93.8\tiny{$\pm 0.2$} & 82.2\tiny{$\pm 0.8$} & 53.1\tiny{$\pm 1.3$} & 25.4\tiny{$\pm 1.2$} & 67.2\tiny{$\pm 0.5$} \\
 & TabICL & 72.5\tiny{$\pm 0.6$} & 89.7\tiny{$\pm 0.5$} & 67.0\tiny{$\pm 0.9$} & 70.3\tiny{$\pm 1.7$} & 89.7\tiny{$\pm 0.3$} & 86.5\tiny{$\pm 0.6$} & 61.3\tiny{$\pm 0.6$} & 67.9\tiny{$\pm 0.7$} & 20.3\tiny{$\pm 0.8$} \\
\cline{1-11}
\multirow[c]{10}{*}{WIDS} & Clean & 77.5 & 77.6 & 79.7 & 75.5 & 72.3 & 49.3 & 80.4 & 70.5 & 75.5 \\
 & RLN & 60.9\tiny{$\pm 0.4$} & 70.1\tiny{$\pm 0.3$} & 70.5\tiny{$\pm 0.3$} & 65.8\tiny{$\pm 0.2$} & 64.7\tiny{$\pm 0.4$} & 46.9\tiny{$\pm 0.1$} & 70.7\tiny{$\pm 0.3$} & 64.5\tiny{$\pm 0.2$} & 67.8\tiny{$\pm 0.4$} \\
 & STG & 68.6\tiny{$\pm 0.5$} & 63.6\tiny{$\pm 0.7$} & 72.4\tiny{$\pm 0.6$} & 69.7\tiny{$\pm 0.2$} & 65.8\tiny{$\pm 0.3$} & 47.2\tiny{$\pm 0.2$} & 73.2\tiny{$\pm 0.5$} & 65.4\tiny{$\pm 0.5$} & 69.6\tiny{$\pm 0.4$} \\
 & TabNet & 76.0\tiny{$\pm 0.1$} & 77.0\tiny{$\pm 0.0$} & 5.5\tiny{$\pm 0.5$} & 69.9\tiny{$\pm 0.2$} & 68.6\tiny{$\pm 0.1$} & 11.9\tiny{$\pm 0.4$} & 67.4\tiny{$\pm 0.2$} & 53.6\tiny{$\pm 0.5$} & 71.8\tiny{$\pm 0.2$} \\
 & TabTr. & 66.5\tiny{$\pm 0.3$} & 72.6\tiny{$\pm 0.2$} & 60.1\tiny{$\pm 0.5$} & 46.0\tiny{$\pm 0.5$} & 62.3\tiny{$\pm 0.4$} & 39.6\tiny{$\pm 0.4$} & 60.8\tiny{$\pm 0.5$} & 54.7\tiny{$\pm 0.6$} & 56.4\tiny{$\pm 0.5$} \\
 & VIME & 69.1\tiny{$\pm 0.2$} & 72.4\tiny{$\pm 0.1$} & 66.9\tiny{$\pm 0.2$} & 65.9\tiny{$\pm 0.2$} & 50.3\tiny{$\pm 0.3$} & 42.6\tiny{$\pm 0.3$} & 66.5\tiny{$\pm 0.3$} & 58.0\tiny{$\pm 0.4$} & 62.8\tiny{$\pm 0.4$} \\
 & RF & 77.5\tiny{$\pm 0.0$} & 77.6\tiny{$\pm 0.1$} & 70.6\tiny{$\pm 1.5$} & 75.1\tiny{$\pm 0.1$} & 71.7\tiny{$\pm 0.1$} & 11.0\tiny{$\pm 0.4$} & 72.3\tiny{$\pm 1.3$} & 62.4\tiny{$\pm 0.8$} & 69.1\tiny{$\pm 1.0$} \\
 & XGB & 76.4\tiny{$\pm 0.4$} & 76.8\tiny{$\pm 0.3$} & 54.7\tiny{$\pm 1.5$} & 73.0\tiny{$\pm 0.3$} & 70.1\tiny{$\pm 0.2$} & 33.6\tiny{$\pm 0.5$} & 35.7\tiny{$\pm 0.7$} & 50.4\tiny{$\pm 1.1$} & 58.8\tiny{$\pm 0.7$} \\
 & TabPFNv2 & 76.0\tiny{$\pm 0.2$} & 76.5\tiny{$\pm 0.2$} & 47.4\tiny{$\pm 1.7$} & 72.3\tiny{$\pm 0.2$} & 69.2\tiny{$\pm 0.2$} & 29.0\tiny{$\pm 1.1$} & 57.3\tiny{$\pm 1.2$} & 30.8\tiny{$\pm 1.0$} & 61.0\tiny{$\pm 1.2$} \\
 & TabICL & 75.3\tiny{$\pm 0.3$} & 75.4\tiny{$\pm 0.1$} & 59.0\tiny{$\pm 1.0$} & 69.9\tiny{$\pm 0.3$} & 68.6\tiny{$\pm 0.3$} & 27.7\tiny{$\pm 0.6$} & 56.7\tiny{$\pm 1.0$} & 46.2\tiny{$\pm 0.8$} & 30.5\tiny{$\pm 0.7$} \\

\bottomrule
\end{tabular}
\end{table*}

%% file: tables/rq3.2_transferability_foundation.tex
% \begin{table*}
% \centering
% \caption{Transferability of foundation models. No results are available for "Same Distribution" and URL as the whole context is always used already.}\label{tab:transferability_foundation}
% \begin{tabular}{llrrrr}
% \toprule
%  &  & \multicolumn{4}{c}{Source} \\
% Dataset & Target & Clean & Subset of context & Same distribution & Exact context  \\
% \midrule
% \multirow[c]{2}{*}{ LCLD} & TabPFNv2 & 55.2 & 44.9\tiny{$\pm 12.2$} & 21.5\tiny{$\pm 2.5$} & 7.5\tiny{$\pm 0.4$} \\
%  & TabICL & 60.5 & 35.5\tiny{$\pm 11.6$} & 28.9\tiny{$\pm 4.2$} & 12.5\tiny{$\pm 0.4$} \\
% \rule{0pt}{2ex} 

% \multirow[c]{2}{*}{URL} & TabPFNv2 & 97.1 & 47.6\tiny{$\pm 4.4$} & - & 26.5\tiny{$\pm 0.9$} \\
%  & TabICL & 96.5 & 52.1\tiny{$\pm 3.6$} & - & 30.7\tiny{$\pm 1.1$} \\

% \rule{0pt}{2ex} 
% \multirow[c]{2}{*}{WIDS} & TabPFNv2 & 68.2 & 39.4\tiny{$\pm 2.6$} & 34.3\tiny{$\pm 1.9$} & 28.9\tiny{$\pm 0.7$} \\
%  & TabICL & 77.0 & 56.8\tiny{$\pm 1.7$} & 48.4\tiny{$\pm 1.8$} & 40.6\tiny{$\pm 1.2$} \\
% \bottomrule
% \end{tabular}
% \end{table*}

\begin{table*}
\centering
\caption{Transferability of foundation models. No results are available for "Same Distribution" and URL as the whole context is always used already.}\label{tab:transferability_foundation}
\normalsize

\begin{tabular}{llrrrr}

\toprule
 &  & \multicolumn{4}{c}{Source} \\
Dataset & Target & Clean & Subset of context & Same distribution & Exact context  \\
\midrule
\multirow[c]{2}{*}{LCLD} & TabPFNv2 & 59.6 & 46.0\tiny{$\pm 6.0$} & 32.6\tiny{$\pm 10.1$} & 9.2\tiny{$\pm 0.4$} \\
 & TabICL & 59.6 & 43.4\tiny{$\pm 11.8$} & 27.9\tiny{$\pm 2.7$} & 10.1\tiny{$\pm 0.4$} \\
\multirow[c]{2}{*}{URL} & TabPFNv2 & 97.0 & 62.2\tiny{$\pm 12.5$} & - & 25.4\tiny{$\pm 1.2$} \\
 & TabICL & 96.7 & 53.5\tiny{$\pm 9.7$} & - & 20.3\tiny{$\pm 0.8$} \\
\multirow[c]{2}{*}{WIDS} & TabPFNv2 & 70.5 & 36.9\tiny{$\pm 6.3$} & 35.4\tiny{$\pm 2.8$} & 30.8\tiny{$\pm 1.0$} \\
 & TabICL & 75.5 & 50.4\tiny{$\pm 5.5$} & 40.3\tiny{$\pm 1.7$} & 30.5\tiny{$\pm 0.7$} \\
\bottomrule
\end{tabular}
\end{table*}

%% file: doc/06_rq3.tex
\section{Adversarial Hardening for Tabular FM}\label{sec:rq3}

\subsection{Tabular FM Defenses}

To defend against evasion attacks, classical defenses involve adversarial fine-tuning (AFT) training, aiming at improving the model robustness by incorporating adversarial examples during the training process. Rather than minimizing the loss on clean inputs $x$, the classifier $h$ is trained to minimize the worst-case loss over a set of allowable perturbations $\Delta$. Formally, the training objective becomes for a tabular foundational model:

\begin{multline}
\label{eq:at}
    \underset{\theta}{\mathrm{min}}~\mathbb{E}_{(X_{test},Y_{test}) \sim D} \\ [\underset{\delta \in \Delta}{\mathrm{max}}~\mathcal{L}(h(X_{test} + \delta| X_{train}, Y_{train}, \theta),  Y_{test})].
\end{multline}

where $\mathcal{L}$ is a loss function (e.g., cross-entropy), and $D$ is the data distribution. Note that the context is fixed here. The inner maximization finds the most adversarial perturbation $\delta$ for the current input, while the outer minimization updates the model parameters $\theta$ to minimize the loss on these adversarially perturbed inputs. In practice, inner maximization is often approximated by using evasion attack methods such as PGD.

However, given the context FM has, an alternative way of doing adversarial training relies on modifying the context $(X_{train}, Y_{train})$. In that case, we do not modify the weights of the models, but rather the context provided to the model along with the inputs $X_{test}$. Intuitively, by making the context itself more robust, we aim to make future predictions robust as well. As such, we define Adversarial In-Context Learning (AICL) as:

\begin{multline}
    \label{eq:aicl}
    \underset{\delta^* \in \Delta^*}{\mathrm{min}}~\mathbb{E}_{(X_{test},Y_{test}) \sim D} \\ [\underset{\delta \in \Delta}{\mathrm{max}}~\mathcal{L}(h(X_{test} + \delta | X_{train} + \delta^*, Y_{train}, \theta),  Y_{test})].
\end{multline}

In that case, perturbations will also be injected into the context, more precisely in $X_{train}$, without modifying the corresponding label, and we mimic the classical adversarial training objectives.

\subsection{Practical Implementation For Tabular FM}\label{sec:alg}

We describe the implementation to optimise the previously introduced objectives in the case of tabular FM.

\paragraph{AFT Implementation} AFT formulation for in-context learning can be similarly applied following Madry adversarial training~\cite{madry2017towards}. The difference stems from the usage of the context. Given a $X_{train}$ dataset (and corresponding labels $Y_{train}$), we split the context into a pseudo-context $\tilde{X}_{train}$ and a pseudo-input $\tilde{X}_{test}$. Then, we run Madry adversarial training~\cite{madry2017towards} by generating adversarial examples over the $\tilde{X}_{test}$ and updating the weights of the underlying model through SGD. This procedure is repeated for several epochs. However, to avoid the model overfitting on a particular context, we proceed as follows 1) between epochs, the $X_{train}$ can be split again, using a different order, 2) inside an epoch, which splits of $X_{train}$ is assigned as $X_{test}$ can be iteratively changed using other splits, so it's not always the same split being used to generate adversarial examples. In the end, the weights are updated, and the context is used as is.

\paragraph{AICL Implementation} In the case of AICL, we follow the same splitting strategy over $X_{train}$ and the epochs that we mentioned in AFT. After each generation of adversarial examples, we update the pseudo-context by replacing the $\tilde{X}_{test}$ used with the new adversarial examples so that, at the next iteration, the context now includes adversarial examples from previous iterations. To keep the perturbation within the $\epsilon$-ball, we generate the adversarial examples from a clean $\tilde{X}_{test}$. As such, only the context is updated while the inputs of the models are kept clean.
% In the end, the procedure will return the best adversarial $\tilde{X}_{train}$ (context), selected based on robust performance over $X{\text{val}}$, along with the default weights.
% % In the end, the procedure will return an adversarial $X_{train}$ (context) and default weights. 
% This approach constitutes a heuristic for the AICL problem formulated in Eq. (\ref{eq:aicl}). 
Intuitively, AICL matches the classical adversarial training procedure with context instead of weights: (i) the model weights are modified via SGD on batches composed of both clean and adversarial data; (ii) The adversarial examples included in train batches are used to approximate the inner maximization in Eq. (\ref{eq:at}); (iii) The adversarial examples are refreshed throughout the training, adapting to the new weights of the model. %Though without any formal convergence guarantees, we decide to use a similar approach in AICL.

Algorithms~\ref{alg:robust_training_aft} and~\ref{alg:robust_training_aicl} give the pseudo-code for the adversarial training procedure outlined in Section~\ref{sec:alg}.

\input{doc/algorithm}

Algorithm~\ref{alg:robust_training_aft} describes the AFT procedure, which adapts traditional adversarial training to tabular FM. It involves training a target model with adversarial training data. However, in the case of a tabular FM, adding the context raises the question of what data will be considered part of the context and what data will be used to improve the model. In particular, the amount of data in the context should be substantial enough to replicate the context at test time. On one hand, keeping $C$ fixed may lead to overfitting on the chosen split. On the other hand, randomly splitting the training data into context and adversarial data may bias the model due to the data distribution. To remediate this issue, we propose making a predetermined number of splits ($n_{split}$) for a given epoch (line 3), which will all be used in turn as the adversarial data (lines 5 - 6). Note that the splits are changed each epoch to avoid having the same data batched together. Then, an evasion attack is applied to generate adversarial examples (line 7), which will be used to improve the model using gradient backpropagation (line 8 - 9). The procedure is repeated so that each split is used to generate adversarial data before changing the splits and restarting the procedure for $n_{epochs}$. In the end, the updated weights $\theta$ are returned.

Algorithm~\ref{alg:robust_training_aicl} describes the AICL procedure. AICL procedure follows mostly the same steps as the AFT procedure we previously described, with one key difference being that we do not update any weights; rather, we update the context given to the model across splits (line 7) and across epochs (line 3). However, since our context becomes increasingly adversarial, we cannot keep applying $\epsilon$ perturbation to new adversarial data. Instead, to remedy this issue, we always apply the perturbation over \textit{clean} data (line 9). This explains why we need to maintain an identical split of both clean and adversarial data (lines 5 - 6). In the end, an updated context is returned (line 16). 

Though this algorithm is a heuristic method (as opposed to gradient-based AFT), we provide a proof of convergence under 5 regularity assumptions in detail in Appendix \ref{sec:app-proof}. 

\begin{theorem}[{Main Convergence Theorem}]
\textcolor{black}{Under the regularity Assumptions A1–A5, AICL converges, i.e.}
\begin{align}
\lim_{t\to\infty}\lVert X'_{train}(t+1)-X'_{train}(t)\rVert & =0. \\
\lim_{t\to\infty}\lVert\nabla\widehat F_{val}(X'_{train}(t))\rVert & =0.
\end{align}
\end{theorem}

We provide in Appendix \ref{sec:app-proof} the full proof of this theorem and detailed comments about the plausibility of each assumption.

\input{tables/rq2.2_robust_foundation}

\input{tables/supp_clean_perf}

\subsection{Experimental Results}\label{sec:rq3-experimental-result}

For tabular FM, we used the previously described procedure that utilises the AdamW~\cite{loshchilov2018decoupled} optimiser with an initial learning rate of 1e-6, both for AFT and AICL. Regarding AICL specifically, we use $5$ splits of the $X_{train}$. We apply both procedures for up to 20 epochs. We used both CAPGD and CAA~\cite{simonetto2024constrained} in adversarial training with $\epsilon = 0.3$ (i.e. our defense epsilon, see Section~\ref{sec:threats}), following standardised benchmarks~\cite{simonetto2024tabularbench}. % For tabular non-FMs models, we use Madry adversarial training with an $\epsilon = 0.3$ and data augmentation. 

 % For AFT, for both tabular FM and traditional models, we retain the weights of the last epoch. For AICL, We kept track of the metric scores obtained on the validation set, and we selected the model with the best validation score in the end.

% \input{tables/rq2.1_robust_dnn}

% \textcolor{red}{We first evaluate the hardening methods applied to all models. The results are given in Table~\ref{tab:rq2_comparison}. In the case of tabular FMs, we focus on AICL adversarial trained with CAA results, as it is the best defence possible (comparing it with the best defence of tabular non-FMs models). Overall, several tabular non-FMs models underperformed on some benchmarks, such as STG on LCLD (10.2\% on CAA), TabTr and VIME on HELOC (2.4\% and 10.2\% respectively on CAA). On the contrary, tabular FMs using AICL prove to be more resilient across the board. While they do not necessarily obtain the best performance on the benchmarks, they obtain competitive results compared to non-FM models based on their clean accuracy and are consistent across benchmarks. \textbf{As such, applying hardening to tabular FM leads to performance improvement, on par with tabular non-FM models, while maintaining stability across benchmarks.}}

The results are reported in Table~\ref{tab:rq2_results}, where we study the difference between the more classical AFT formulation and the newly proposed AICL for hardening. For both adversarial training procedures, we used both CAA and the weaker CAPGD attack. First, we note that AFT does not necessarily lead to improvement over no-adversarial training (Original), particularly when CAPGD is used. As such, acting on the weights of tabular FMs models might not be the most effective approach to hardening. On the contrary, AICL outperforms AFT and Original in all cases. In several cases, AICL using CAPGD proved to be even better than AFT using CAA, which is a more powerful attack. 

\textbf{Conclusion:} \textit{as it yields better improvements than traditional adversarial hardening, in-context adversarial learning offers interesting avenues to develop new defences for tabular FM.}

To further extend the analysis, we present the clean performance of the models across all tabular metrics on the entire test set, as shown in Table~\ref{tab:clean_perf}, obtained while using the different hardening methods. 

\textbf{Conclusion:} \textit{Overall, hardening methods have little impact on the accuracy of clean data compared to no hardening.}

%% file: doc/algorithm.tex
\begin{algorithm}[h]
\caption{Adversarial Fine-Tuning (AFT) algorithm. For each $X$ there is a $Y$ containing its associated correct labels.}\label{alg:robust_training_aft}
\SetAlgoLined
\SetKwComment{Comment}{/* }{ */}

\KwData{$X_{train}$, $n_{epochs}$, $\theta$ weights of the transformer $f$, $n_{split}$, Attack, $\epsilon$}

$X'_{train} \gets X_{train}$\;
\For{$i \in [1,n_{epochs}]$}{
  $X'_1, ..., X'_{n_{split}} \gets $make\_split($X'_{train}, n_{split}$, $seed=i$)\;
  \For{$j \in [1, n_{split}]$}{
    $\widetilde{X}_{train} \gets \{X'_{k \neq j} \forall k \in [1, n_{split}] \}$\;
    $\widetilde{X}_{val} \gets X'_j$\; 
    
    $\widetilde{X}_{adv} \gets$ Attack($\widetilde{X}_{train}, \widetilde{X}_{val}, \epsilon, \theta, ...)$\;
    $\widetilde{Y}_{pred} \gets f(context=[\widetilde{X}_{train}, \widetilde{Y}_{train}], test=\widetilde{X}_{adv}, \theta)$\;
      $\theta \gets \theta - \alpha \nabla_{\theta}Loss(\widetilde{Y}_{val}, \widetilde{Y}_{pred})$; \Comment{Update model weights}
    
  }
}
\textbf{return} $\theta$\;
\end{algorithm}

\begin{algorithm}[h]
\caption{Adversarial In-Context Learning (AICL) algorithm. For each $X$ there is a $Y$ containing its associated correct labels.}\label{alg:robust_training_aicl}
\SetAlgoLined
\SetKwComment{Comment}{/* }{ */}

\KwData{$X_{train}$, $n_{epochs}$, $\theta$ weights of the transformer $f$, $n_{split}$, Attack, $\epsilon$}

$X'_{train} \gets X_{train}$\;

\Comment{Initialize a mask $M$ that decides which samples to be perturbed}
$M \gets$ create\_mask($X_{train}$)\;

\For{$i \in [1,n_{epochs}]$}{
  $M_1, ..., M_{n_{split}} \gets $make\_split($M, n_{split}$, $seed=i$)\;
  % $I_1, ..., I_{n_{split}} \gets $make\_split($\#M, n_{split}$, $seed=i$)\;
  %\Comment{Second identical split to have the X test always clean}
  
  $X'_1, ..., X'_{n_{split}} \gets $make\_split($X'_{train}, n_{split}$, $seed=i$)\;
  $X_1, ..., X_{n_{split}} \gets $make\_split($X_{train}, n_{split}$, $seed=i$)\;
  %\Comment{Second identical split to have the X test always clean}
  %$X_1, ..., X_{n_{split}} \gets $make\_split($X_{train}, n_{split}$, $seed=i$)\;
  
  \For{$j \in [1, n_{split}]$}{
    $\widetilde{X}_{train} \gets \{X'_{k \neq j} \forall k \in [1, n_{split}] \}$\;
    %\Comment{Here, it should be clean!}
    %$\widetilde{X}_{val} \gets X_j$\; 
    
    %\Comment{Select only the masked subset of $\widetilde{X}_{val}$ for perturbation}
    \Comment{Select only the masked subset of $X_{j}$ for perturbation}
    $\widetilde{X}_{val}^{mask} \gets X_{j}[M{j}]$\;
    $\widetilde{X}_{val}^{clean} \gets X_{j}[\neg M{j}]$\;
    
    $\widetilde{X}_{adv} \gets$ Attack($\widetilde{X}_{train}, \widetilde{X}_{val}^{mask}, \epsilon, \theta, ...)$\;
    
    \Comment{Concatenate perturbed and unperturbed samples back together}
    $X'_j \gets \widetilde{X}_{adv} \cup \widetilde{X}_{val}^{clean}$\;
  }
  
  $X'_{train} \gets X'_1, ..., X'_{n_{split}}$; \Comment{Update context}
}
\textbf{return} $X'_{train}$\;
\end{algorithm}

% \begin{algorithm}[h]
% \caption{Adversarial In-Context Learning (AICL) algorithm. For each $X$ there is a $Y$ containing its associated correct labels.}\label{alg:robust_training_aicl}
% \SetAlgoLined
% \SetKwComment{Comment}{/* }{ */}

% \KwData{$X_{train}$, $n_{epochs}$, $\theta$ weights of the transformer $f$, $n_{split}$, Attack, $\epsilon$}

% $X'_{train} \gets X_{train}$\;
% \For{$i \in [1,n_{epochs}]$}{
%   $X'_1, ..., X'_{n_{split}} \gets $make\_split($X'_{train}, n_{split}$, $seed=i$)\;
%   % $\hat{\epsilon} \gets \epsilon$ \;
%   % \If{is\_in\_context\_adversarial\_training}{
%   %       \Comment{For in context, make sure adversarial examples remain in $\epsilon$-ball}
%   %       $\hat{\epsilon} \gets \frac{6\epsilon}{\pi^2i^2}$\; 
%   % }
%   \Comment{Need to have a second identical split to have the X test always clean}
%   $X_1, ..., X_{n_{split}} \gets $make\_split($X_{train}, n_{split}$, $seed=i$)\;
  
%   \For{$j \in [1, n_{split}]$}{
%     $\widetilde{X}_{train} \gets \{X'_{k \neq j} \forall k \in [1, n_{split}] \}$\;
%     \Comment{Here, it should be clean!}
%     $\widetilde{X}_{val} \gets X_j$\; 
    
%     $\widetilde{X}_{adv} \gets$ Attack($\widetilde{X}_{train}, \widetilde{X}_{test}, \epsilon, \theta, ...)$\;
%     $X'_j \gets \widetilde{X}_{adv}$; \Comment{Update local context}

%   }
  
%   $X'_{train} \gets X'_1, ..., X'_{n_{split}}$; \Comment{Update context}
% }
% \textbf{return} $X'_{train}$\;
% \end{algorithm}

%% file: tables/rq2.2_robust_foundation.tex
\begin{table*}[t]
\centering
\caption{Robust accuracy on Clean data and adversarial CAPGD/CAA attacks using AICL and AFT hardening methods}\label{tab:rq2_results}

\small
\resizebox{0.75\textwidth}{!}{%
\begin{tabular}{llrrrrrrrrr}
\toprule
Dataset &  & \multicolumn{3}{c}{LCLD} & \multicolumn{3}{c}{URL} & \multicolumn{3}{c}{WIDS} \\%& \multicolumn{3}{c}{HELOC} \\
\cmidrule(r){3-5} \cmidrule(r){6-8} \cmidrule(r){9-11} % \cmidrule(r){12-14}
 Model & Training & Clean & CAPGD & CAA & Clean & CAPGD & CAA & Clean & CAPGD & CAA \\%& Clean & CAPGD & CAA \\

\midrule

\multirow[c]{5}{*}{TabPFN2} 

% & Original (met)& 59.6 & 17.3 \tiny{$\pm 0.1$} & 8.5 \tiny{$\pm 0.0$} & 97.0 & 42.4 \tiny{$\pm 0.2$} & 23.7 \tiny{$\pm 0.3$} & 70.5 & 33.1 \tiny{$\pm 0.3$} & 26.6 \tiny{$\pm 0.3$} \\
% %& 73.4 & 0.5 \tiny{$\pm 0.1$} & 0.0 \tiny{$\pm 0.0$} \\

& Original & 59.7 & 22.4 \tiny{$\pm 0.1$} & 9.0 \tiny{$\pm 0.2$} & 97.1 & 47.6 \tiny{$\pm 0.2$} & 25.2 \tiny{$\pm 0.5$} & 70.7 & 49.2 \tiny{$\pm 0.4$} & 30.8 \tiny{$\pm 0.4$} \\%& 73.5 & 5.0 \tiny{$\pm 0.6$} & 0.3 \tiny{$\pm 0.1$} \\

& AICL \tiny{CAA} & 60.4 & 49.3 \tiny{$\pm 0.3$} & 41.2 \tiny{$\pm 0.0$} & 96.2 & 96.1 \tiny{$\pm 0.0$} & 69.4 \tiny{$\pm 0.2$} & 70.7 & 65.7 \tiny{$\pm 0.3$} & 51.5 \tiny{$\pm 0.2$}\\% & 73.7 & 73.0 \tiny{$\pm 0.2$} & 35.4 \tiny{$\pm 0.5$} \\

& AICL \tiny{CAPGD}& 55.8 & 42.0 \tiny{$\pm 0.5$} & 14.7 \tiny{$\pm 0.2$} & 95.6 & 94.7 \tiny{$\pm 0.1$} & 25.9 \tiny{$\pm 0.2$} & 70.7 & 63.7 \tiny{$\pm 0.2$} & 42.5 \tiny{$\pm 0.4$} \\%& 68.8 & 67.9 \tiny{$\pm 0.2$} & 7.4 \tiny{$\pm 0.3$} \\

& AFT \tiny{CAA} & 62.7 & 20.6 \tiny{$\pm 0.4$} & 12.0 \tiny{$\pm 0.1$} & 97.2 & 51.1 \tiny{$\pm 0.2$} & 31.2 \tiny{$\pm 0.3$} & 73.3 & 38.2 \tiny{$\pm 0.3$} & 31.1 \tiny{$\pm 0.5$} \\%& 74.5 & 0.3 \tiny{$\pm 0.1$} & 0.1 \tiny{$\pm 0.1$} \\

& AFT \tiny{CPGD}& 61.6 & 22.3 \tiny{$\pm 0.3$} & 12.8 \tiny{$\pm 0.1$} & 97.1 & 53.2 \tiny{$\pm 0.4$} & 31.8 \tiny{$\pm 0.1$} & 68.7 & 34.2 \tiny{$\pm 0.3$} & 25.7 \tiny{$\pm 0.2$} \\%& 78.3 & 0.7 \tiny{$\pm 0.1$} & 0.2 \tiny{$\pm 0.0$} \\

\rule{0pt}{1ex} \\

\multirow[c]{5}{*}{TabICL} 

% & Original (met) & 59.6 & 18.3 \tiny{$\pm 0.3$} & 9.9 \tiny{$\pm 0.2$} & 96.7 & 82.8 \tiny{$\pm 0.2$} & 20.4 \tiny{$\pm 0.3$} & 75.5 & 38.4 \tiny{$\pm 0.1$} & 29.9 \tiny{$\pm 0.1$} & 70.9 & 11.9 \\ 
% %\tiny{$\pm 0.2$} & 10.9 \tiny{$\pm 0.1$} \\

& Original & 59.6 & 18.5 \tiny{$\pm 0.4$} & 10.2 \tiny{$\pm 0.1$} & 96.7 & 82.9 \tiny{$\pm 0.2$} & 20.2 \tiny{$\pm 0.3$} & 75.5 & 38.2 \tiny{$\pm 0.3$} & 30.5 \tiny{$\pm 0.3$} \\%& 70.9 & 11.9 \tiny{$\pm 0.1$} & 11.0 \tiny{$\pm 0.1$} \\

& AICL \tiny{CAA} & 60.3 & 51.2 \tiny{$\pm 0.3$} & 39.6 \tiny{$\pm 0.2$} & 97.3 & 97.0 \tiny{$\pm 0.0$} & 75.6 \tiny{$\pm 0.2$} & 76.2 & 60.0 \tiny{$\pm 0.2$} & 54.7 \tiny{$\pm 0.2$} \\%& 72.5 & 72.4 \tiny{$\pm 0.0$} & 40.8 \tiny{$\pm 0.2$}\\

& AICL \tiny{CAPGD}& 58.8 & 47.1 \tiny{$\pm 0.2$} & 20.8 \tiny{$\pm 0.2$} & 96.4 & 95.8 \tiny{$\pm 0.1$} & 31.0 \tiny{$\pm 0.4$} & 75.5 & 55.5 \tiny{$\pm 0.2$} & 50.3 \tiny{$\pm 0.2$} \\%& 66.5 & 66.5 \tiny{$\pm 0.0$} & 10.8 \tiny{$\pm 0.2$} \\

& AFT \tiny{CAA} & 54.4 & 22.6 \tiny{$\pm 0.1$} & 20.4 \tiny{$\pm 0.2$} & 96.6 & 80.3 \tiny{$\pm 0.2$} & 22.6 \tiny{$\pm 0.2$} & 71.6 & 44.1 \tiny{$\pm 0.3$} & 32.2 \tiny{$\pm 0.4$} \\%& 73.9 & 17.3 \tiny{$\pm 0.2$} & 7.2 \tiny{$\pm 0.2$} \\

& AFT \tiny{CPGD} & 59.2 & 18.4 \tiny{$\pm 0.2$} & 4.3 \tiny{$\pm 0.1$} & 96.7 & 84.8 \tiny{$\pm 0.3$} & 16.8 \tiny{$\pm 0.1$} & 54.6 & 35.2 \tiny{$\pm 0.4$} & 14.8 \tiny{$\pm 0.3$} \\%& 75.5 & 3.4 \tiny{$\pm 0.3$} & 0.1 \tiny{$\pm 0.1$} \\

\bottomrule
\end{tabular}%
}
\end{table*}

%% file: tables/supp_clean_perf.tex
\begin{table*}

\caption{Metrics of clean data test set on Tabular FMs with varying training method.}\label{tab:clean_perf}
\centering
\small
\setlength\tabcolsep{3pt}
\begin{tabular}{lllrrrrrr}
\toprule
 &  & Metric & AUROC & MCC & F1 & Accuracy & Recall & Precision \\
Dataset & Arch. & Training&  &  &  &  &  &  \\
\midrule
\multirow[c]{6}{*}{LCLD} & 
\multirow[c]{4}{*}{TabPFNv2} 
& Original & 71.2 & 25.4 & 43.3 & 68.1 & 60.1 & 33.9 \\
 &  & AICL \tiny{CAA}  & 71.0 & 25.0 & 43.1 & 67.5 & 60.8 & 33.4 \\
 &  & AFT \tiny{CAA} & 71.2 & 25.2 & 43.3 & 66.9 & 62.5 & 33.2 \\

\cline{2-9}
 & \multirow[c]{3}{*}{TabICL} 
 & Original & 71.4 & 25.5 & 43.4 & 68.1 & 60.5 & 33.9 \\
 &  & AICL \tiny{CAA} & 71.4 & 25.4 & 43.4 & 68.0 & 60.6 & 33.8 \\
 &  & AFT \tiny{CAA} & 71.1 & 25.1 & 42.8 & 70.3 & 54.8 & 35.1 \\

\cline{1-9}
\multirow[c]{6}{*}{URL} & 
\multirow[c]{4}{*}{TabPFNv2} 
& Original & 99.2 & 92.8 & 96.4 & 96.4 & 96.8 & 96.1 \\
 &  & AICL \tiny{CAA} & 98.9 & 90.6 & 95.3 & 95.3 & 96.2 & 94.5 \\
 &  & AFT \tiny{CAA} & 99.2 & 92.7 & 96.4 & 96.4 & 97.0 & 95.8 \\

\cline{2-9}
 & \multirow[c]{3}{*}{TabICL} 
 & Original & 99.4 & 92.8 & 96.4 & 96.4 & 96.9 & 96.0 \\
 &  & AICL \tiny{CAA} & 99.3 & 91.4 & 95.7 & 95.7 & 97.2 & 94.3 \\
 &  & AFT \tiny{CAA} & 99.4 & 93.1 & 96.6 & 96.5 & 96.8 & 96.3 \\

\cline{1-9}
\multirow[c]{6}{*}{WIDS} & 
\multirow[c]{4}{*}{TabPFNv2} 
& Original & 88.2 & 43.0 & 47.0 & 85.8 & 70.5 & 35.3 \\
 &  & AICL \tiny{CAA} & 88.0 & 42.5 & 46.4 & 85.4 & 70.7 & 34.6 \\
 &  & AFT \tiny{CAA} & 88.2 & 42.3 & 45.9 & 84.5 & 73.3 & 33.4 \\

\cline{2-9}
 & \multirow[c]{3}{*}{TabICL} 
 & Original & 88.3 & 42.0 & 45.2 & 83.6 & 75.5 & 32.2 \\
 &  & AICL \tiny{CAA} & 87.9 & 39.4 & 42.4 & 81.5 & 76.2 & 29.4 \\
 &  & AFT \tiny{CAA} & 88.4 & 44.0 & 47.8 & 86.0 & 71.6 & 35.8 \\

%  \cline{1-9}
% \multirow[c]{6}{*}{HELOC} & 
% \multirow[c]{4}{*}{TabPFNv2} 
% & Original & 78.4 & 43.5 & 70.6 & 71.7 & 73.4 & 68.0 \\
%  &  & AICL \tiny{CAA} & 77.1 & 41.1 & 69.7 & 70.3 & 73.7 & 66.1 \\
%  &  & AFT \tiny{CAA} & 78.3 & 43.8 & 70.9 & 71.8 & 74.5 & 67.7 \\

% \cline{2-9}
%  & \multirow[c]{3}{*}{TabICL} 
%  & Original & 77.7 & 41.3 & 69.2 & 70.7 & 70.9 & 67.4 \\
%  &  & AICL \tiny{CAA}  & 76.7 & 40.4 & 69.1 & 70.1 & 72.5 & 66.1 \\
%  &  & AFT \tiny{CAA} & 77.5 & 41.9 & 70.1 & 70.8 & 73.9 & 66.6 \\

\bottomrule
\end{tabular}
\end{table*}

%% file: doc/06_bis_discussion.tex
\section{Discussion on Tabular FM hardening}

In our study, we demonstrated the weakness of tabular FM to evasion attacks, both obtained from scratch and through transfer using either tabular FM or non-FM. AICL proved to be an effective alternative for tabular FM hardening compared to the more classical AFT procedure. \textcolor{black}{Intuitively, as tabular-FMs rely more on their context than weights when tackling a new task, we believe this is why re-training with adversarial examples with in-context perturbations yields better performances compared to examples with weights perturbations. On top of that, AICL could be deployable in practice as traditional adversarial training would; the computation time (Table \ref{tab:app-cost} in Appendix \ref{sec:app-protocol}) that AICL has a similar computation time as AFT, not accounting for the lack of optimization of our implementation.} To prolong the reflection, we compare the hardening obtained via AICL to the strongest defence on non-FM models (Madry). Results are given in Table~\ref{tab:rq2_comparison} below.

\input{tables/rq2.1_robust_dnn}

Overall, several tabular non-FMs models underperformed on some benchmarks (e.g. STG on LCLD - 13.4\% on CAA). On the contrary, tabular FMs using AICL (with CAA) prove to be more resilient across the board. While they do not necessarily obtain the best performance on the benchmarks, they obtain competitive results compared to non-FM models based on their clean accuracy and are consistent across benchmarks. Moreover, using the weaker defence based on CAPGD, shows that the defence still increases the robustness, albeit less than what can be obtained on most non-FM models.

While one argues that using CAA-light can lead to a false sense of security on CAA \cite{carlini2019evaluatingadversarialrobustness}, we showed that AICL - CAPGD is better than AFT - CPGD. Given the difference of settings between tabular FM and non-FM models (e.g. in-context learn, training on a synthetic dataset etc.), this last one can be considered a close translation of Madry's defence for FM models. For a similar time budget, AICL outperforms AFT. In practice, AICL even converge faster than AFT.

\textbf{Conclusion:} \textit{As such, applying hardening to tabular FM leads to performance improvement, on par with tabular non-FM models, while maintaining stability across benchmarks.}

%% file: tables/rq2.1_robust_dnn.tex
\begin{table*}
\centering

\caption{Performance using the robustification with Madry adversarial training for specialised models and AICL for tabular foundational models. Robust accuracy on Clean data and advesarial CAPGD and CAA. 
} 

\label{tab:rq2_comparison}
\small
\resizebox{0.75\textwidth}{!}{
\begin{tabular}{lrrrrrrrrr}
\toprule
Dataset & \multicolumn{3}{c}{LCLD} & \multicolumn{3}{c}{URL} & \multicolumn{3}{c}{WIDS}  \\%\multicolumn{3}{c}{HELOC} \\
\cmidrule(r){2-4} \cmidrule(r){5-7} \cmidrule(r){8-10} %\cmidrule(r){11-13}

Model & Clean & CAPGD & CAA & Clean & CAPGD & CAA & Clean & CAPGD & CAA \\ %& Clean & CAPGD & CAA \\
\midrule

% RLN & 75.1 & 55.9\tiny{$\pm 0.2$} & 55.5\tiny{$\pm 0.2$} & 93.9 & 86.3\tiny{$\pm 0.1$} & 64.3\tiny{$\pm 0.5$} & 77.5 & 73.3\tiny{$\pm 0.2$} & 70.9\tiny{$\pm 0.4$}  & & \\
RLN & 70.3 & 64.3 \tiny{$\pm 0.1$} & 63.0 \tiny{$\pm 0.1$} & 95.2 & 81.3 \tiny{$\pm 0.1$} & 52.3 \tiny{$\pm 0.2$} & 78.0 & 67.6 \tiny{$\pm 0.2$} & 66.6 \tiny{$\pm 0.2$} \\ %& 90.1 & 24.7 \tiny{$\pm 0.1$} & 24.5 \tiny{$\pm 0.1$} \\

% STG & 82.6 & 81.5\tiny{$\pm 0.1$} & 81.5\tiny{$\pm 0.1$} & 94.3 & 91.6\tiny{$\pm 0.0$} & 89.8\tiny{$\pm 0.1$} & 85.3 & 74.8\tiny{$\pm 0.3$} & 73.6\tiny{$\pm 0.3$}  & & \\
STG & 16.7 & 13.4 \tiny{$\pm 0.1$} & 13.4 \tiny{$\pm 0.1$} & 94.3 & 91.6 \tiny{$\pm 0.1$} & 89.8 \tiny{$\pm 0.1$} & 62.6 & 46.5 \tiny{$\pm 0.2$} & 45.1 \tiny{$\pm 0.1$} \\ %& 88.8 & 15.5 \tiny{$\pm 0.1$} & 15.1 \tiny{$\pm 0.1$} \\

% TabNet & 67.5 & 7.3\tiny{$\pm 0.4$} & 0.2\tiny{$\pm 0.1$} & 90.1 & 90.0\tiny{$\pm 0.0$} & 89.9\tiny{$\pm 0.0$} & 79.7 & 7.9\tiny{$\pm 0.3$} & 5.5\tiny{$\pm 0.3$}  & & \\
TabNet & 0 & 0 \tiny{$\pm 0.0$} & 0 \tiny{$\pm 0.0$} & 99.5 & 93.8 \tiny{$\pm 0.0$} & 91.8 \tiny{$\pm 0.1$} & 98.4 & 58.8 \tiny{$\pm 0.2$} & 58.5 \tiny{$\pm 0.1$} \\ %& 100.0 & 100.0 \tiny{$\pm 0.0$}& 100.0 \tiny{$\pm 0.0$}\\

% TabTr. & 80.5 & 79.6\tiny{$\pm 0.1$} & 79.5\tiny{$\pm 0.0$} & 93.0 & 88.4\tiny{$\pm 0.1$} & 65.0\tiny{$\pm 0.5$} & 78.1 & 69.0\tiny{$\pm 0.4$} & 68.1\tiny{$\pm 0.2$}  & & \\
TabTr. & 74.5 & 70.5 \tiny{$\pm 0.1$} & 69.9 \tiny{$\pm 0.1$} & 93.9 & 81.0 \tiny{$\pm 0.0$} & 52.9 \tiny{$\pm 0.3$} & 77.3 & 66.3 \tiny{$\pm 0.2$} & 65.2 \tiny{$\pm 0.1$} \\ %& 70.3 & 3.1 \tiny{$\pm 0.1$} & 2.4  \tiny{$\pm 0.0$} \\

% VIME & 80.5 & 77.3\tiny{$\pm 0.1$} & 77.2\tiny{$\pm 0.1$} & 91.9 & 82.3\tiny{$\pm 0.1$} & 71.8\tiny{$\pm 0.8$} & 72.1 & 63.1\tiny{$\pm 0.3$} & 62.1\tiny{$\pm 0.5$}  & & \\
VIME & 65.1 & 12.8 \tiny{$\pm 0.2$} & 10.2 \tiny{$\pm 0.1$} & 93.4 & 78.8 \tiny{$\pm 0.1$} & 67.8 \tiny{$\pm 0.2$} & 72.1 & 53.0 \tiny{$\pm 0.1$} & 52.3 \tiny{$\pm 0.1$} \\ %& 74.9 & 10.3 \tiny{$\pm 0.1$} & 10.2 \tiny{$\pm 0.0$} \\

\tablespace
TabPFNv2 \tiny{CAA} & 60.4 & 49.3 \tiny{$\pm 0.3$} & 41.2 \tiny{$\pm 0.0$} & 96.2 & 96.1 \tiny{$\pm 0.0$} & 69.4 \tiny{$\pm 0.2$} & 70.7 & 65.7 \tiny{$\pm 0.3$} & 51.5 \tiny{$\pm 0.2$} \\ %& 73.7 & 73.0 \tiny{$\pm 0.2$} & 35.4 \tiny{$\pm 0.5$} \\

TabPFNv2 \tiny{CAPGD}& 55.8 & 42.0 \tiny{$\pm 0.5$} & 14.7 \tiny{$\pm 0.2$} & 95.6 & 94.7 \tiny{$\pm 0.1$} & 25.9 \tiny{$\pm 0.2$} & 70.7 & 63.7 \tiny{$\pm 0.2$} & 42.5 \tiny{$\pm 0.4$} \\ %& 68.8 & 67.9 \tiny{$\pm 0.2$} & 7.4 \tiny{$\pm 0.3$} \\

TabICL \tiny{CAA} & 60.3 & 51.2 \tiny{$\pm 0.3$} & 39.6 \tiny{$\pm 0.2$} & 97.3 & 97.0 \tiny{$\pm 0.0$} & 75.6 \tiny{$\pm 0.2$} & 76.2 & 60.0 \tiny{$\pm 0.2$} & 54.7 \tiny{$\pm 0.2$} \\ %& 72.5 & 72.4 \tiny{$\pm 0.0$} & 40.8 \tiny{$\pm 0.2$}\\

TabICL \tiny{CAPGD}& 58.8 & 47.1 \tiny{$\pm 0.2$} & 20.8 \tiny{$\pm 0.2$} & 96.4 & 95.8 \tiny{$\pm 0.1$} & 31.0 \tiny{$\pm 0.4$} & 75.5 & 55.5 \tiny{$\pm 0.2$} & 50.3 \tiny{$\pm 0.2$} \\ %& 66.5 & 66.5 \tiny{$\pm 0.0$} & 10.8 \tiny{$\pm 0.2$} \\
\bottomrule
\end{tabular}}
\end{table*}

%% file: doc/07_limitations.tex
\section{Limitations and Opportunities}\label{sec:limitations}

Our study is the first formalisation and evaluation of the constrained adversarial robustness of tabular FM in context. It naturally paves the way to multiple new directions:

\textbf{Largest pool of models and datasets:} %Multiple new models proposed improvements to the in-context paradigm of TabPFN. 
We could validate our insights on a more diverse set of models. However, to our best knowledge, none of the available improvements to TabPFN would impact the intrinsic vulnerabilities of tabular FM to adversarial attacks. Similarly, we show consistent insights in vulnerability, transferability, and robustification on three datasets of varying size and complexity, \textcolor{black}{and additional datasets should not significantly affect our findings}.

\textbf{Exhaustive evaluation of available robustification mechanisms:} There is an extended literature of defence mechanisms with variable effectiveness and requirements. We focus our evaluation on the most established mechanism, adversarial training, and propose a new robustification paradigm that optimises the context itself. We believe existing defences are orthogonal to AICL, and could be combined into stronger defences. In particular, AICL and AFT could be theoretically combined since they focus on different parts of the model. Yet, this raises several challenges precisely because of this. As such, we leave such a study to future work.

\textbf{Formal guarantees and interpretation:} \textcolor{black}{Similarly to works related to in-context learning in NLP \cite{zhao2024universalvulnerabilitieslargelanguage, abad2024context} and Computer Vision \cite{wei2023jailbroken, zhao2023evaluating}, we provide an extensive empirical evaluation of the phenomena.} We give proofs that AICL follows a similar paradigm as traditional adversarial training, and thus we believe it can be assessed through the different theories behind adversarial vulnerability, like the non-robust features hypothesis or the flatness of the loss surface hypothesis. We argue that this new paradigm offers opportunities to understand adversarial robustness and its underlying mechanisms better. 

\textbf{Additional adversarial scenarios:} We set our threat model to be the most favourable for the attacker: white-box settings with test-time data knowledge. While this might not be representative of the majority of practical attack scenarios, by exploring the most difficult scenario in terms of defence, we effectively obtain an upper bound on the capacity of the proposed defence. As such, we believe our approach would prove more effective in a more constrained threat model (e.g. black-box settings). \textcolor{black}{Our approach could be extended to categorical problems. From a security point of view, our threat model would not be changed: a multi-class untargeted threat model boils down to the binary case (correct class or not), and a targeted threat model (misclassification in a chosen target class) is straightforward as well as it would follow the principles of equivalent threat models. Nonetheless, studying the robustness of tabular-FMs is an interesting future work.}

The vulnerability we have uncovered in white-box and transferable evasion attacks can also manifest itself for backdoor and poisoning attacks, and we believe that AICL could help improve the robustness to these threat models. Nevertheless, such attacks would require a different threat model, out of the scope of this study, and so we leave to future work the study of backdoor and poisoning attacks against tabular FMs. Finally, we do not study the robustness of tabular FMs against real-world attacks. \textcolor{black}{Finally, for our datasets, CAA and similar attacks can be considered real-world attacks because problem objects are directly represented with numerical features. Yet, in other cases (e.g. Malware), this might not be the case, and so the robustness of tabular-FMs might be even lower. Studying such real-world attacks is an interesting additional scenario, which nonetheless would require creating them (i.e. very costly) and is out of scope of the paper.}

%% file: doc/08_conclusion.tex
\section{Conclusion}

We explored the robustness of tabular FM to evasion attacks. We demonstrate that popular tabular FM models are significantly more vulnerable than specialised models to state-of-the-art tabular evasion attacks and are viable surrogates for transferable attacks. Our results for three datasets and application domains motivate the development of novel defences tailored to tabular in-context learners.

To this end, we offer a new formulation of adversarial hardening for in-context learners that updates the context (Adversarial In-Context Learning). We evaluated AICL and an adapted formulation of the classical adversarial training (AFT) hardening and confirmed that AICL leads to better robustness. While AICL on tabular FM is more stable than Madry's defence applied on non-FM models, the most robust non-FM models outperform FM models. Our definition and implementation of AICL opens major research directions in the robustness evaluation of tabular FM, and paves the way for advanced and effective defences against security attacks for in-context tabular FM. On a larger scale, our research contributes to enabling the trustworthy deployment of tabular FM in real-world industry applications.

%% file: doc/80_appendix.tex
%Please refer to the appendices in the supplementary material for complete references.

\input{doc/appendices/related-work}
\input{doc/appendices/experimental-protocol}

\input{doc/appendices/rep}

\input{doc/appendices/rq1}
\input{doc/appendices/rq3}
\input{doc/appendices/proof}
\input{doc/appendices/budget_experiments_classical}
% \newpage
% \input{doc/appendices/rebuttal}

%% file: doc/appendices/related-work.tex
\section{Complementary related work}\label{sec:app-related-comp}

\subsection{Tabular Foundation Models}

Multiple variants of TabPFN have been proposed in the literature; we only present below general-purpose TabPFN variants, and defer the readers to this survey~\cite{jiang2025tabularsurvey} for additional tabular foundation models, especially tuned for specific tasks and domains (cyber-security, time-series, ...).

\paragraph{TabPFN~\cite{hollmann2022tabpfn}}
represents a groundbreaking approach to tabular data classification, introducing the first transformer-based foundation model specifically designed for small tabular datasets. The model performs in-context learning (ICL) by accepting both training and test samples as set-valued input and yielding predictions for the entire test set in a single forward pass, requiring no hyperparameter tuning. TabPFN is trained offline once on synthetic datasets drawn from a prior that incorporates ideas from causal reasoning, entailing a large space of structural causal models with a preference for simple structures. On datasets with up to 1,000 training points, 100 numerical features, and 10 classes, TabPFN demonstrates remarkable efficiency, outperforming boosted trees and matching complex AutoML systems with up to 230× speedup on CPU and 5,700× speedup on GPU

\paragraph{TabForestPFN~\cite{denbreejen2024tabforestpfn}}
extends the TabPFN framework by introducing fine-tuning capabilities and a novel forest dataset generator that creates datasets with complex decision boundaries. The key innovation lies in pretraining ICL-transformers on synthetic datasets that, while unrealistic in nature, possess intricate decision boundaries that enhance the model's ability to handle complex classification tasks. The model demonstrates that fine-tuning enables ICL-transformers to create complex decision boundaries, a property that regular neural networks lack. By combining both the original TabPFN synthetic dataset generator and the new forest generator, TabForestPFN achieves excellent fine-tuning performance while maintaining good zero-shot capabilities.

\paragraph{TabDPT (Tabular Discriminative Pre-trained Transformer)~\cite{ma2024tabdpt}}
addresses the scalability limitations of existing tabular ICL approaches by training tabular-specific ICL-based architectures on real data using self-supervised learning and retrieval techniques. Unlike TabPFN's synthetic-only pretraining, TabDPT combines the advantages of transformer-based architectures with real-world data patterns, enabling effective in-context learning across diverse tabular tasks without task-specific fine-tuning. The model demonstrates strong scaling properties as both model size and available data increase, achieving state-of-the-art performance on CC18 classification and CTR23 regression benchmarks. TabDPT's approach of incorporating retrieval mechanisms allows the model to reference relevant data points during prediction, significantly enhancing its adaptability to new tasks and datasets.

\paragraph{APT (Adversarially Pre-trained Transformer)~\cite{wu2025zeroshotmetalearningtabularprediction}}
introduces a novel framework for zero-shot meta-learning on tabular data through adversarial pretraining using synthetic data agents. The model's core innovation lies in its use of adversarial synthetic data generators that continuously evolve their data distributions based on the model's performance, creating increasingly challenging datasets that enhance the model's robustness and generalization capabilities. APT incorporates a mixture block architecture to effectively handle variations in classification tasks, particularly addressing the challenge of varying numbers of classes across different datasets. This adversarial training approach enables APT to be more adaptable and generalizable compared to traditional approaches that rely on fixed synthetic data distributions, extending the capabilities of existing models like TabPFN to handle unseen datasets more effectively.

\paragraph{TabICL (Tabular In-Context Learning)~\cite{qu2025tabicl}}
represents a significant advancement in scalable tabular foundation models, designed to handle datasets with up to 500,000 samples and 500 features through efficient in-context learning. The model employs a unique two-stage architecture: first creating dense vector embeddings of table rows using a distribution-aware column-wise embedding strategy with Set Transformers to handle permutation-invariance, followed by row-wise interaction modeling through transformer architectures. TabICL's column-then-row attention mechanism reduces computational complexity while enabling effective processing of data with varying numbers of features. Pretrained on synthetic datasets with up to 60,000 samples, TabICL significantly expands the scalability boundaries of ICL for tabular data, integrating tree-based structural causal models using XGBoost and employing curriculum learning through progressively increasing synthetic dataset sizes.

\paragraph{TabPFNv2~\cite{hollmann2025tabpfn}}
represents a significant advancement over its predecessor, introducing specialized feature tokenization to better handle data heterogeneity. The model transforms each input feature into a k-dimensional vector using a shared mapping, with random position encoding vectors added to differentiate attributes across datasets. This innovation enables TabPFN v2 to handle larger datasets with up to 10,000 samples, 500 dimensions, and expanded class limits. Comprehensive evaluations on over 300 tabular datasets demonstrate TabPFN v2's exceptional generalizability on small- to medium-scale datasets, achieving a 25.06\% probability of achieving the best accuracy/RMSE across diverse benchmark tasks. The model's success stems from its ability to unify heterogeneous datasets into fixed-dimensional representations through randomized feature tokens, enabling more effective training and inference without requiring dataset-specific tuning.

\subsection{Related Work on Adversarial Robustness in In-Context Learning}

While in-context learning (ICL) has been extensively studied in the context of natural language processing, its application to tabular data is a burgeoning area of research. Recent studies have begun to explore the adversarial robustness of ICL in tabular settings.

\cite{anwar2024adversarialrobustnessincontextlearning} investigate the susceptibility of single-layer linear transformers, which emulate gradient descent in-context, to \emph{hijacking attacks}. They demonstrate that perturbing a single example in the in-context training set can coerce the model into producing arbitrary predictions. While these attacks are effective on linear transformers, they do not readily transfer to more complex architectures like GPT-2. However, such models can still be compromised using gradient-based adversarial techniques. The study also finds that adversarial training, even when applied solely during fine-tuning, enhances robustness against these attacks. Interestingly, training against weaker attack models can sometimes confer resilience to stronger adversarial strategies.

Complementing this, Yu et al.~\cite{yu2024evaluating} examine the robustness of retrieval-augmented ICL methods. Their findings indicate that while retrieval mechanisms can bolster defense against test sample perturbations, they may inadvertently increase vulnerability to attacks targeting the demonstrations themselves. To address this, they propose DARD, a training-free adversarial defense that enriches the example pool with adversarially perturbed samples, leading to improved performance and robustness.

In the realm of natural language processing, He et al.~\cite{he2024using} explore the integration of natural language explanations (NLEs) into ICL. Their approach involves augmenting prompts with human-generated NLEs, which the model then uses to generate further explanations. This method yields over a 6\% improvement in robustness across various adversarial datasets, including HANS, ISCS, and ANLI, demonstrating the efficacy of NLEs in enhancing model resilience.

In contrast, our study uniquely addresses adversarial robustness in structured, tabular data, introducing an adversarial in-context learning approach that improves robustness by modifying only context data, thus avoiding parameter updates.

%% file: doc/appendices/experimental-protocol.tex
\section{Additional experimental protocol details}
\label{sec:app-protocol}

\subsection{Datasets}
\label{subsec:dataset_app}

\begin{table*}[ht]
  \centering
  \caption{The datasets evaluated in the empirical study, with the class imbalance of each dataset (Balance in \%).}
  \small
  \label{tab:data_extended}
  \begin{tabular}{l|llll}
    \toprule
     & \multicolumn{4}{c}{Properties} \\
     Dataset & Task & Size & \# Features & Balance \\
    \midrule
    LCLD~\cite{lcld} &  Credit Scoring &  1 220 092 & 28 & 80/20 \\
    URL~\cite{hannousse2021towards} & Phishing URL detection    & 11 430 & 63  & 50/50 \\
    WIDS~\cite{wids} & ICU patient survival & 91 713 & 186& 91.4/8.6 \\
    \bottomrule
  \end{tabular}
\end{table*}

Our dataset design followed the same protocol as the TabularBench benchmark\cite{simonetto2024tabularbench}.
We present in Table~\ref{tab:data_extended} the attributes of our datasets.

\paragraph{Credit Scoring - LCLD} (license: CC0: Public Domain) This dataset is derived from the publicly accessible Lending Club Loan Data.\footnote{https://www.kaggle.com/wordsforthewise/lending-club} This dataset contains 151 features, where each row is a loan processed by the Lending Club paltform. Nevertheless, some of these accepted loans are charged off and remain unpaid. Our objective is to predict, at the time of the request, whether the borrower will repay the loan or if it will be charged off. We processed dataset from Tabularbench has 47 input features and one target feature. We split the dataset using random sampling stratified by the target class, resulting in a training set of 915K examples and a testing set of 305K examples. Both sets are unbalanced, with only 20\% of loans being charged off (class 1). For each feature in this dataset, we define boundary constraints based on the extreme values observed in the training set. We consider the 19 features under the control of the Lending Club as immutable. We reuse the same 10 relationship constraints (3 linear and 7 non-linear) defined in Tabularbench.

\paragraph{URL Phishing - ISCX-URL2016} (license CC BY 4.0) Phishing attacks are frequently used to carry out cyber fraud or identity theft. Such attacks usually include a URL that resembles a genuine one (such as a favorite online shopping site), yet it leads the user to a deceitful website that requests personal or financial details. The features extracted from the URL include the number of special substrings such as ``www'', ``\&'', ``,'', ``\$'', ``and'', the length of the URL, the port, the presence of a brand in the domain, subdomain, or path, and the inclusion of ``http'' or ``https''. External service-based features include the Google index, page rank, and the domain's presence in DNS records. The full list of features is available in the reproduction package. The dataset contains 5715 legitimate and 5715 malicious URLs. We use 75\% of the dataset for training and validation, and the remaining 25\% for testing and adversarial generation. The dataset has 14 relational constraints between the URL features. Among these, 7 are linear constraints (e.g., the length of the hostname is less than or equal to the length of the URL) and 7 are Boolean constraints of the form $if\ a > 0 \ then\ b > 0$ (e.g., if the number of ``http'' $>$ 0, then the number of slashes ``/'' $>$ 0).

\paragraph{WiDS} (license: PhysioNet Restricted Health Data License 1.5.0 )\footnote{https://physionet.org/content/widsdatathon2020/view-license/1.0.0/} ~\cite{wids} dataset contains medical data on the survival of patients admitted to the ICU. The objective is to predict whether a patient will survive or die based on biological features (e.g., for triage). This highly unbalanced dataset has 30 linear relational constraints.

\paragraph{HELOC}~\cite{heloc_dataset_update} dataset contains financial and credit history of homeowners. The objective is to predict if the applicants will make timely payments over two years, classifying them as "Good" or "Bad". We add this dataset as an extra experiment.

\subsection{Model Architectures}
\label{sec:app_model_arch}
\begin{table*}
\centering
  \caption{The model architectures of our study.}
  \label{tab:models}
  \small
  \begin{tabular}{lll}
  
    \toprule
    Family & Model & Hyperparameters\\
    \midrule

    % Foundational  & TabPFNv2, TabICL & \begin{tabular}[c]{@{}l@{}}$n\_estimators$, $softmax\_temperature$,\\ $balance\_probabilities$ \end{tabular}  \\
    
    Transformer & TabTransformer  & \begin{tabular}[c]{@{}l@{}}$hidden\_dim$, $n\_layers$,\\ $learning\_rate$, $norm$, $\theta$\end{tabular}  \\ 
    
    Transformer & TabNet  & \begin{tabular}[c]{@{}l@{}}$n\_d$, $n\_steps$,\\ $\gamma$, $cat\_emb\_dim$, $n\_independent$, \\ 
    $n\_shared$, $momentum$, $mask\_type$ \end{tabular}  \\
    
    Regularization & RLN & \begin{tabular}[c]{@{}l@{}}$hidden\_dim$, $depth$, \\ $heads$, $weight\_decay$, \\ $learning\_rate$, $dropout$\end{tabular} \\
    
    Regularization & STG & \begin{tabular}[c]{@{}l@{}} $hidden\_dims$, $learning\_rate$, $lam$\end{tabular} \\ \\
    
    Encoding & VIME  &  $p_m$, $\alpha$, $K$, $\beta$\\
        
    % Tree-based & XGB & \begin{tabular}[c]{@{}l@{}}$max\_depth$, $alpha$, $llambda$,\\ $eta$ \end{tabular}  \\
    % Tree-based & RF & \begin{tabular}[c]{@{}l@{}}$n\_estimators$, $max\_depth$ \end{tabular}  \\

  \bottomrule
\end{tabular}
\end{table*}

Table~\ref{tab:models} provides an overview of the family, model architecture, and hyperparameters adjusted during the training of our models.
\paragraph{TabTransformer} is a transformer-based model~\cite{huang2020tabtransformer}.Self-attention is used to transform categorical features into a contextual embedding, which the paper claims improves the model's ability to handle noisy inputs.
\paragraph{TabNet} is another model based on transformers~\cite{arik2021tabnet}. It uses several sub-networks arranged in sequence. At each decision step, it implements sequential attention to determine which features should be focused on. The results from each step are aggregated by TabNet to produce the final decision.
\paragraph{RLN} or Regularization Learning Networks~\cite{shavitt2018regularization} uses an effective hyperparameter tuning to minimize counterfactual loss. The approach involves training a regularization coefficient for the weights on the neural network to decrease sensitivity, which results in highly sparse networks.
\paragraph{STG} or Stochastic Gates~\cite{icml2020_5085} leverages stochastic gates for feature selectionThe technique is based on a probabilistic relaxation of the $l_0$ norm of features i.e., the count of selected features.
\paragraph{VIME} or Value Imputation for Mask Estimation~\cite{yoon2020vime} uses self-supervised and semi-supervised learning encoders and predictors.

\subsection{Hardware and Computation Resources}

We run our experiments on a server node with 40 cores / 80 threads and 503GB and four NVIDIA L40S with 46GB of VRAM.
We parallelize 4 tasks, each running on its dedicated GPU.

CPU and RAM resources are negligible compared to GPU and VRAM resource to attack tabular foundation models. Hence, we estimate the required ressource in GPU hours. 
The complete attack of a foundation model with CAA takes on average one hour. MOEVA accounts for more than 90\% of the required resources. 
The attack time of non-foundation models time is negligible. 
For transferability experiments, we save the attacks generated with the source model, and replay them on the target model. Therefore, the execution time of this experiment is negligible if the white-box setting has already been calculated (e.g. Table~\ref{tab:transferability_inter_arch}).

We repeated all experiments with five random seeds.
We report below the marginal cost of adding each experiment in the paper.
The experiments in Table~\ref{tab:rq1_results} require 12 GPU hours.
Table~\ref{tab:transferability_inter_arch} is free.
Table~\ref{tab:transferability_foundation} requires an extra 60 GPU hours.
%Figure~\ref{fig:tabpfn_comparison} is a visualization of the results of Table~\ref{tab:transferability_inter_arch} and does not require additional computation.
Table~\ref{tab:rq2_results} requires an extra 60 GPU hours while Table~\ref{tab:rq2_comparison} is another presentation of the results for the foundation model part. 

In the appendices, Figures~\ref{fig:eps} and ~\ref{fig:n_steps} require an extra 60 GPU hour each.
Figure~\ref{fig:n_gen}, because of MOEVA's changed budget requires 75 GPU hours. 
Figure~\ref{fig:subset} has a marginal cost of 30 GPU hours.
Transferability attacks of Section~\ref{sec:rq2} are free.
Training clean models takes seconds on foundation model architecture. However, the robustification attempt takes one hour on average for URL and LCLD and two hours WIDS, for a total of 24 GPU hours. 

The total cost of the study is therefore 459 GPU hours or 7.65 GPU days.

We estimate that the total cost of the full research project, accounting for failed runs and unsuccessful experiments, is approximately three times this amount.

\begin{table}[htbp]
    \caption{Computation cost in seconds of AICL$_{CAPGD}$ across the different models}
    \label{tab:app-cost}
    \centering
    \begin{tabular}{lllll}
\toprule
Model & Training & LCLD & URL & WIDS \\
\midrule
TabPFNv2 & AFT & 7223 & 7671 & 20441 \\
TabPFNv2 & AICL & 7155 & 7355 & 20260 \\
TabICL & AFT & 4533 & 4519 & 8806 \\
TabICL & AICL & 4510 & 4376 & 8601 \\
\bottomrule
    \end{tabular}
\end{table}

%% file: doc/appendices/rep.tex
\subsection{Reproduction Package and Availability}\label{sec:reproduction}

% The source code, datasets, and pre-trained models to reproduce the experiments of this paper are available at \url{https://figshare.com/projects/TabFM/249944}. The source code will be available publicly upon acceptance under the MIT license or similar.

The source code to reproduce the experiments of this paper is available at \url{https://github.com/serval-uni-lu/tabularbench_tabfm_aicl.git}.

%% file: doc/appendices/rq3.tex
\section{Complementary resources of RQ3}
\label{sec:app-rq3}

\subsection{Additional Hardening Experiments Including HELOC Dataset}

We give in the following tables: the robust/clean accuracy using AICL/AFT in Table \ref{tab:app-rq2_results}, the clean metrics of the test set with different adversarial training for tabular FM in Table \ref{tab:app-clean_perf} and the comparison with non-FM models in Table \ref{tab:app-rq2_comparison}.\footnote{TabNet returned accuracy of 0\% on LCLD (always 0), and always predicted 1 on HELOC (hence the 100\%). For completeness sake, we leave it in the table, but the performances are not representative.} Results on HELOC are consistent with what was obtained on other datasets.

\begin{table*}[t]
\centering
\caption{Robust accuracy and clean accuracy using AICL and AFT on tabular FMs models}\label{tab:app-rq2_results}

\small
\resizebox{\textwidth}{!}{%
\begin{tabular}{llrrrrrrrrrrrr}
\toprule
Dataset &  & \multicolumn{3}{c}{LCLD} & \multicolumn{3}{c}{URL} & \multicolumn{3}{c}{WIDS} & \multicolumn{3}{c}{HELOC} \\
\cmidrule(r){3-5} \cmidrule(r){6-8} \cmidrule(r){9-11} \cmidrule(r){12-14}
 Model & Training & Clean & CAPGD & CAA & Clean & CAPGD & CAA & Clean & CAPGD & CAA & Clean & CAPGD & CAA \\

\midrule

\multirow[c]{5}{*}{TabPFNv2} 

% & Originalor & 59.6 & 17.3 \tiny{$\pm 0.1$} & 8.5 \tiny{$\pm 0.0$} & 97.0 & 42.4 \tiny{$\pm 0.2$} & 23.7 \tiny{$\pm 0.3$} & 70.5 & 33.1 \tiny{$\pm 0.3$} & 26.6 \tiny{$\pm 0.3$} & 73.4 & 0.5 \tiny{$\pm 0.1$} & 0.0 \tiny{$\pm 0.0$} \\

% & Original (Meteor) & 59.6 & 17.3 \tiny{$\pm 0.1$} & 8.5 \tiny{$\pm 0.0$} & 97.0 & 42.4 \tiny{$\pm 0.2$} & 23.7 \tiny{$\pm 0.3$} & 70.5 & 33.1 \tiny{$\pm 0.3$} & 26.6 \tiny{$\pm 0.3$} & 73.4 & 0.5 \tiny{$\pm 0.1$} & 0.0 \tiny{$\pm 0.0$} \\

% & Original (Precision)& 59.7 & 22.4 \tiny{$\pm 0.1$} & 9.0 \tiny{$\pm 0.2$} & 97.1 & 47.6 \tiny{$\pm 0.2$} & 25.2 \tiny{$\pm 0.5$} & 70.7 & 49.2 \tiny{$\pm 0.4$} & 30.8 \tiny{$\pm 0.4$} & 73.5 & 5.0 \tiny{$\pm 0.6$} & 0.3 \tiny{$\pm 0.1$} \\

& Original & 59.7 & 22.4 \tiny{$\pm 0.1$} & 9.0 \tiny{$\pm 0.2$} & 97.1 & 47.6 \tiny{$\pm 0.2$} & 25.2 \tiny{$\pm 0.5$} & 70.7 & 49.2 \tiny{$\pm 0.4$} & 30.8 \tiny{$\pm 0.4$} & 73.5 & 5.0 \tiny{$\pm 0.6$} & 0.3 \tiny{$\pm 0.1$} \\

& AICL \tiny{CAA} & 60.4 & 49.3 \tiny{$\pm 0.3$} & 41.2 \tiny{$\pm 0.0$} & 96.2 & 96.1 \tiny{$\pm 0.0$} & 69.4 \tiny{$\pm 0.2$} & 70.7 & 65.7 \tiny{$\pm 0.3$} & 51.5 \tiny{$\pm 0.2$} & 73.7 & 73.0 \tiny{$\pm 0.2$} & 35.4 \tiny{$\pm 0.5$} \\

& AICL \tiny{CAPGD}& 55.8 & 42.0 \tiny{$\pm 0.5$} & 14.7 \tiny{$\pm 0.2$} & 95.6 & 94.7 \tiny{$\pm 0.1$} & 25.9 \tiny{$\pm 0.2$} & 70.7 & 63.7 \tiny{$\pm 0.2$} & 42.5 \tiny{$\pm 0.4$} & 68.8 & 67.9 \tiny{$\pm 0.2$} & 7.4 \tiny{$\pm 0.3$} \\

& AFT \tiny{CAA} & 62.7 & 20.6 \tiny{$\pm 0.4$} & 12.0 \tiny{$\pm 0.1$} & 97.2 & 51.1 \tiny{$\pm 0.2$} & 31.2 \tiny{$\pm 0.3$} & 73.3 & 38.2 \tiny{$\pm 0.3$} & 31.1 \tiny{$\pm 0.5$} & 74.5 & 0.3 \tiny{$\pm 0.1$} & 0.1 \tiny{$\pm 0.1$} \\

& AFT \tiny{CPGD}& 61.6 & 22.3 \tiny{$\pm 0.3$} & 12.8 \tiny{$\pm 0.1$} & 97.1 & 53.2 \tiny{$\pm 0.4$} & 31.8 \tiny{$\pm 0.1$} & 68.7 & 34.2 \tiny{$\pm 0.3$} & 25.7 \tiny{$\pm 0.2$} & 78.3 & 0.7 \tiny{$\pm 0.1$} & 0.2 \tiny{$\pm 0.0$} \\

\rule{0pt}{1ex} \\

\multirow[c]{5}{*}{TabICL} 

& Original & 59.6 & 18.3 \tiny{$\pm 0.3$} & 9.9 \tiny{$\pm 0.2$} & 96.7 & 82.8 \tiny{$\pm 0.2$} & 20.4 \tiny{$\pm 0.3$} & 75.5 & 38.4 \tiny{$\pm 0.1$} & 29.9 \tiny{$\pm 0.1$} & 70.9 & 11.9 \tiny{$\pm 0.2$} & 10.9 \tiny{$\pm 0.1$} \\

% & Original (Meteor) & 59.6 & 18.3 \tiny{$\pm 0.3$} & 9.9 \tiny{$\pm 0.2$} & 96.7 & 82.8 \tiny{$\pm 0.2$} & 20.4 \tiny{$\pm 0.3$} & 75.5 & 38.4 \tiny{$\pm 0.1$} & 29.9 \tiny{$\pm 0.1$} & 70.9 & 11.9 \tiny{$\pm 0.2$} & 10.9 \tiny{$\pm 0.1$} \\

% & Original (Precision)& 59.6 & 18.5 \tiny{$\pm 0.4$} & 10.2 \tiny{$\pm 0.1$} & 96.7 & 82.9 \tiny{$\pm 0.2$} & 20.2 \tiny{$\pm 0.3$} & 75.5 & 38.2 \tiny{$\pm 0.3$} & 30.5 \tiny{$\pm 0.3$} & 70.9 & 11.9 \tiny{$\pm 0.1$} & 11.0 \tiny{$\pm 0.1$} \\

& AICL \tiny{CAA} & 60.3 & 51.2 \tiny{$\pm 0.3$} & 39.6 \tiny{$\pm 0.2$} & 97.3 & 97.0 \tiny{$\pm 0.0$} & 75.6 \tiny{$\pm 0.2$} & 76.2 & 60.0 \tiny{$\pm 0.2$} & 54.7 \tiny{$\pm 0.2$} & 72.5 & 72.4 \tiny{$\pm 0.0$} & 40.8 \tiny{$\pm 0.2$}\\

& AICL \tiny{CAPGD}& 58.8 & 47.1 \tiny{$\pm 0.2$} & 20.8 \tiny{$\pm 0.2$} & 96.4 & 95.8 \tiny{$\pm 0.1$} & 31.0 \tiny{$\pm 0.4$} & 75.5 & 55.5 \tiny{$\pm 0.2$} & 50.3 \tiny{$\pm 0.2$} & 66.5 & 66.5 \tiny{$\pm 0.0$} & 10.8 \tiny{$\pm 0.2$} \\

& AFT \tiny{CAA} & 54.4 & 22.6 \tiny{$\pm 0.1$} & 20.4 \tiny{$\pm 0.2$} & 96.6 & 80.3 \tiny{$\pm 0.2$} & 22.6 \tiny{$\pm 0.2$} & 71.6 & 44.1 \tiny{$\pm 0.3$} & 32.2 \tiny{$\pm 0.4$} & 73.9 & 17.3 \tiny{$\pm 0.2$} & 7.2 \tiny{$\pm 0.2$} \\

& AFT \tiny{CPGD} & 59.2 & 18.4 \tiny{$\pm 0.2$} & 4.3 \tiny{$\pm 0.1$} & 96.7 & 84.8 \tiny{$\pm 0.3$} & 16.8 \tiny{$\pm 0.1$} & 54.6 & 35.2 \tiny{$\pm 0.4$} & 14.8 \tiny{$\pm 0.3$} & 75.5 & 3.4 \tiny{$\pm 0.3$} & 0.1 \tiny{$\pm 0.1$} \\

\bottomrule
\end{tabular}%
}
\end{table*}

\begin{table*}

\caption{Clean metrics of test set on foundation model with varying training method.}\label{tab:app-clean_perf}
\centering
\small
\setlength\tabcolsep{3pt}
\begin{tabular}{lllrrrrrr}
\toprule
 &  & Metric & AUROC & MCC & F1 & Accuracy & Recall & Precision \\
Dataset & Arch. & Training&  &  &  &  &  &  \\
\midrule
\multirow[c]{6}{*}{LCLD} & 
\multirow[c]{4}{*}{TabPFNv2} 
& Original & 71.2 & 25.4 & 43.3 & 68.1 & 60.1 & 33.9 \\
 &  & AICL \tiny{CAA}  & 71.0 & 25.0 & 43.1 & 67.5 & 60.8 & 33.4 \\
 &  & AFT \tiny{CAA} & 71.2 & 25.2 & 43.3 & 66.9 & 62.5 & 33.2 \\

\cline{2-9}
 & \multirow[c]{3}{*}{TabICL} 
 & Original & 71.4 & 25.5 & 43.4 & 68.1 & 60.5 & 33.9 \\
 &  & AICL \tiny{CAA} & 71.4 & 25.4 & 43.4 & 68.0 & 60.6 & 33.8 \\
 &  & AFT \tiny{CAA} & 71.1 & 25.1 & 42.8 & 70.3 & 54.8 & 35.1 \\

\cline{1-9}
\multirow[c]{6}{*}{URL} & 
\multirow[c]{4}{*}{TabPFNv2} 
& Original & 99.2 & 92.8 & 96.4 & 96.4 & 96.8 & 96.1 \\
 &  & AICL \tiny{CAA} & 98.9 & 90.6 & 95.3 & 95.3 & 96.2 & 94.5 \\
 &  & AFT \tiny{CAA} & 99.2 & 92.7 & 96.4 & 96.4 & 97.0 & 95.8 \\

\cline{2-9}
 & \multirow[c]{3}{*}{TabICL} 
 & Original & 99.4 & 92.8 & 96.4 & 96.4 & 96.9 & 96.0 \\
 &  & ICL \tiny{CAA} & 99.3 & 91.4 & 95.7 & 95.7 & 97.2 & 94.3 \\
 &  & AFT \tiny{CAA} & 99.4 & 93.1 & 96.6 & 96.5 & 96.8 & 96.3 \\

\cline{1-9}
\multirow[c]{6}{*}{WIDS} & 
\multirow[c]{4}{*}{TabPFNv2} 
& Original & 88.2 & 43.0 & 47.0 & 85.8 & 70.5 & 35.3 \\
 &  & AICL \tiny{CAA} & 88.0 & 42.5 & 46.4 & 85.4 & 70.7 & 34.6 \\
 &  & AFT \tiny{CAA} & 88.2 & 42.3 & 45.9 & 84.5 & 73.3 & 33.4 \\

\cline{2-9}
 & \multirow[c]{3}{*}{TabICL} 
 & Original & 88.3 & 42.0 & 45.2 & 83.6 & 75.5 & 32.2 \\
 &  & ICL \tiny{CAA} & 87.9 & 39.4 & 42.4 & 81.5 & 76.2 & 29.4 \\
 &  & AFT \tiny{CAA} & 88.4 & 44.0 & 47.8 & 86.0 & 71.6 & 35.8 \\

 \cline{1-9}
\multirow[c]{6}{*}{HELOC} & 
\multirow[c]{4}{*}{TabPFNv2} 
& Original & 78.4 & 43.5 & 70.6 & 71.7 & 73.4 & 68.0 \\
 &  & AICL \tiny{CAA} & 77.1 & 41.1 & 69.7 & 70.3 & 73.7 & 66.1 \\
 &  & AFT \tiny{CAA} & 78.3 & 43.8 & 70.9 & 71.8 & 74.5 & 67.7 \\

\cline{2-9}
 & \multirow[c]{3}{*}{TabICL} 
 & Original & 77.7 & 41.3 & 69.2 & 70.7 & 70.9 & 67.4 \\
 &  & AICL \tiny{CAA}  & 76.7 & 40.4 & 69.1 & 70.1 & 72.5 & 66.1 \\
 &  & AFT \tiny{CAA} & 77.5 & 41.9 & 70.1 & 70.8 & 73.9 & 66.6 \\

\bottomrule
\end{tabular}
\end{table*}

\begin{table*}

\caption{\textcolor{black}{Clean metrics of test set on default classical models.}}\label{tab:app-clean_perf-classical}
\centering
\small
\setlength\tabcolsep{3pt}
\begin{tabular}{lllrrrrrr}
\toprule
 &  & Metric & AUROC & MCC & F1 & Accuracy & Recall & Precision \\
Dataset & Arch. & Training&  &  &  &  &  &  \\
\midrule

\multirow[c]{7}{*}{LCLD}
&\multirow[c]{1}{*}{RLN} & Original & 71.7 & 25.3 & 43.5 & 63.8 & 68.7 & 31.8 \\ 
& \multirow[c]{1}{*}{STG} & Original & 70.7 & 24.4 & 42.9 & 64.2 & 66.2 & 31.7 \\ 
& \multirow[c]{1}{*}{TabNet} & Original & 72.1 & 26.0 & 43.9 & 65.3 & 67.1 & 32.6 \\ 
& \multirow[c]{1}{*}{TabTr} & Original & 71.5 & 25.2 & 43.4 & 63.0 & 70.0 & 31.4 \\ 
& \multirow[c]{1}{*}{VIME} & Original & 71.2 & 24.9 & 43.2 & 64.1 & 67.4 & 31.8 \\
& \multirow[c]{1}{*}{RF} & Original & 71.6 & 25.6 & 43.6 & 66.1 & 64.7 & 32.9 \\
& \multirow[c]{1}{*}{XGB} & Original & 72.1 & 26.1 & 44.0 & 64.7 & 68.2 & 32.4 \\

\cline{1-9}
\multirow[c]{7}{*}{URL}
& \multirow[c]{1}{*}{RLN} & Original & 98.4 & 89.1 & 94.5 & 94.5 & 94.6 & 94.5 \\ 
& \multirow[c]{1}{*}{STG} & Original & 97.3 & 83.9 & 92.1 & 92.0 & 93.4 & 90.8 \\ 
& \multirow[c]{1}{*}{TabNet} & Original & 98.6 & 89.2 & 94.5 & 94.6 & 93.7 & 95.4 \\ 
& \multirow[c]{1}{*}{TabTr} & Original & 98.1 & 88.0 & 94.0 & 94.0 & 93.7 & 94.3 \\ 
& \multirow[c]{1}{*}{VIME} & Original & 97.4 & 85.6 & 92.8 & 92.8 & 92.7 & 92.9 \\
& \multirow[c]{1}{*}{RF} & Original & 99.1 & 90.3 & 95.2 & 95.1 & 95.8 & 94.6 \\
& \multirow[c]{1}{*}{XGB} & Original & 99.4 & 93.8 & 96.9 & 96.9 & 97.5 & 96.4 \\

\cline{1-9}
\multirow[c]{7}{*}{WIDS} 
& \multirow[c]{1}{*}{RLN} & Original & 87.0 & 37.7 & 40.6 & 79.7 & 77.5 & 27.5 \\ 
& \multirow[c]{1}{*}{STG} & Original & 86.7 & 36.3 & 39.1 & 78.4 & 77.6 & 26.2 \\ 
& \multirow[c]{1}{*}{TabNet} & Original & 87.1 & 36.7 & 39.1 & 77.8 & 79.7 & 25.9 \\ 
& \multirow[c]{1}{*}{TabTr} & Original & 87.4 & 38.5 & 41.7 & 81.1 & 75.5 & 28.8 \\ 
& \multirow[c]{1}{*}{VIME} & Original & 86.7 & 38.7 & 42.5 & 82.5 & 72.3 & 30.1 \\
& \multirow[c]{1}{*}{RF} & Original & 87.8 & 43.8 & 48.8 & 90.8 & 49.3 & 48.4 \\
& \multirow[c]{1}{*}{XGB} & Original & 88.8 & 40.5 & 42.7 & 80.7 & 80.4 & 29.1 \\

\cline{1-9}
\multirow[c]{7}{*}{HELOC}
& \multirow[c]{1}{*}{RLN} & Original & 76.9 & 41.6 & 68.8 & 70.9 & 69.4 & 68.3 \\ 
& \multirow[c]{1}{*}{STG} & Original & 72.7 & 26.1 & 52.4 & 63.4 & 43.4 & 65.9 \\ 
& \multirow[c]{1}{*}{TabNet} & Original & 76.5 & 42.1 & 69.9 & 71.0 & 72.9 & 67.2 \\ 
& \multirow[c]{1}{*}{TabTr} & Original & 76.1 & 39.9 & 68.7 & 69.9 & 71.4 & 66.2 \\ 
& \multirow[c]{1}{*}{VIME} & Original & 76.4 & 39.3 & 68.5 & 69.6 & 71.4 & 65.8 \\
& \multirow[c]{1}{*}{RF} & Original & 77.6 & 42.6 & 70.0 & 71.3 & 72.5 & 67.7 \\
& \multirow[c]{1}{*}{XGB} & Original & 77.9 & 42.4 & 69.2 & 71.3 & 69.6 & 68.8 \\

\bottomrule
\end{tabular}
\end{table*}

\begin{table*}
\centering

\caption{Performance using the robustification with Madry adversarial training for specialised models and AICL for tabular foundational models. Robust accuracy CAPGD and CAA. The Clean column corresponds to the accuracy of the model on the subset of clean samples that we attack. 
} 

\label{tab:app-rq2_comparison}
\small
\resizebox{\textwidth}{!}{
\begin{tabular}{lrrrrrrrrrrrr}
\toprule
Dataset & \multicolumn{3}{c}{LCLD} & \multicolumn{3}{c}{URL} & \multicolumn{3}{c}{WIDS} & \multicolumn{3}{c}{HELOC} \\
\cmidrule(r){2-4} \cmidrule(r){5-7} \cmidrule(r){8-10} \cmidrule(r){11-13}

Model & Clean & CAPGD & CAA & Clean & CAPGD & CAA & Clean & CAPGD & CAA & Clean & CAPGD & CAA \\
\midrule

% RLN & 75.1 & 55.9\tiny{$\pm 0.2$} & 55.5\tiny{$\pm 0.2$} & 93.9 & 86.3\tiny{$\pm 0.1$} & 64.3\tiny{$\pm 0.5$} & 77.5 & 73.3\tiny{$\pm 0.2$} & 70.9\tiny{$\pm 0.4$}  & & \\
RLN & 70.3 & 64.3 \tiny{$\pm 0.1$} & 63.0 \tiny{$\pm 0.1$} & 95.2 & 81.3 \tiny{$\pm 0.1$} & 52.3 \tiny{$\pm 0.2$} & 78.0 & 67.6 \tiny{$\pm 0.2$} & 66.6 \tiny{$\pm 0.2$} & 90.1 & 24.7 \tiny{$\pm 0.1$} & 24.5 \tiny{$\pm 0.1$} \\

% STG & 82.6 & 81.5\tiny{$\pm 0.1$} & 81.5\tiny{$\pm 0.1$} & 94.3 & 91.6\tiny{$\pm 0.0$} & 89.8\tiny{$\pm 0.1$} & 85.3 & 74.8\tiny{$\pm 0.3$} & 73.6\tiny{$\pm 0.3$}  & & \\
STG & 16.7 & 13.4 \tiny{$\pm 0.1$} & 13.4 \tiny{$\pm 0.1$} & 94.3 & 91.6 & 89.8 \tiny{$\pm 0.1$} & 62.6 & 46.5 \tiny{$\pm 0.2$} & 45.1 \tiny{$\pm 0.1$} & 88.8 & 15.5 \tiny{$\pm 0.1$} & 15.1 \tiny{$\pm 0.1$} \\

% TabNet & 67.5 & 7.3\tiny{$\pm 0.4$} & 0.2\tiny{$\pm 0.1$} & 90.1 & 90.0\tiny{$\pm 0.0$} & 89.9\tiny{$\pm 0.0$} & 79.7 & 7.9\tiny{$\pm 0.3$} & 5.5\tiny{$\pm 0.3$}  & & \\
TabNet & 0 \tiny{$\pm 0.0$} & 0 \tiny{$\pm 0.0$} & 0 \tiny{$\pm 0.0$} & 99.5 & 93.8 \tiny{$\pm 0.0$} & 91.8 \tiny{$\pm 0.1$} & 98.4 & 58.8 \tiny{$\pm 0.2$} & 58.5 \tiny{$\pm 0.1$} & 100.0 & 100.0 \tiny{$\pm 0.0$}& 100.0 \tiny{$\pm 0.0$}\\

% TabTr. & 80.5 & 79.6\tiny{$\pm 0.1$} & 79.5\tiny{$\pm 0.0$} & 93.0 & 88.4\tiny{$\pm 0.1$} & 65.0\tiny{$\pm 0.5$} & 78.1 & 69.0\tiny{$\pm 0.4$} & 68.1\tiny{$\pm 0.2$}  & & \\
TabTr. & 74.5 & 70.5 \tiny{$\pm 0.1$} & 69.9 \tiny{$\pm 0.1$} & 93.9 & 81.0 \tiny{$\pm 0.0$} & 52.9 \tiny{$\pm 0.3$} & 77.3 & 66.3 \tiny{$\pm 0.2$} & 65.2 \tiny{$\pm 0.1$} & 70.3 & 3.1 \tiny{$\pm 0.1$} & 2.4  \tiny{$\pm 0.0$} \\

% VIME & 80.5 & 77.3\tiny{$\pm 0.1$} & 77.2\tiny{$\pm 0.1$} & 91.9 & 82.3\tiny{$\pm 0.1$} & 71.8\tiny{$\pm 0.8$} & 72.1 & 63.1\tiny{$\pm 0.3$} & 62.1\tiny{$\pm 0.5$}  & & \\
VIME & 65.1 & 12.8 \tiny{$\pm 0.2$} & 10.2 \tiny{$\pm 0.1$} & 93.4 & 78.8 \tiny{$\pm 0.1$} & 67.8 \tiny{$\pm 0.2$} & 72.1 & 53.0 \tiny{$\pm 0.1$} & 52.3 \tiny{$\pm 0.1$} & 74.9 & 10.3 \tiny{$\pm 0.1$} & 10.2 \tiny{$\pm 0.0$} \\

\tablespace
TabPFNv2 \tiny{CAA} & 60.4 & 49.3 \tiny{$\pm 0.3$} & 41.2 \tiny{$\pm 0.0$} & 96.2 & 96.1 \tiny{$\pm 0.0$} & 69.4 \tiny{$\pm 0.2$} & 70.7 & 65.7 \tiny{$\pm 0.3$} & 51.5 \tiny{$\pm 0.2$} & 73.7 & 73.0 \tiny{$\pm 0.2$} & 35.4 \tiny{$\pm 0.5$} \\

TabPFNv2 \tiny{CAPGD}& 55.8 & 42.0 \tiny{$\pm 0.5$} & 14.7 \tiny{$\pm 0.2$} & 95.6 & 94.7 \tiny{$\pm 0.1$} & 25.9 \tiny{$\pm 0.2$} & 70.7 & 63.7 \tiny{$\pm 0.2$} & 42.5 \tiny{$\pm 0.4$} & 68.8 & 67.9 \tiny{$\pm 0.2$} & 7.4 \tiny{$\pm 0.3$} \\
TabICL \tiny{CAA} & 60.3 & 51.2 \tiny{$\pm 0.3$} & 39.6 \tiny{$\pm 0.2$} & 97.3 & 97.0 \tiny{$\pm 0.0$} & 75.6 \tiny{$\pm 0.2$} & 76.2 & 60.0 \tiny{$\pm 0.2$} & 54.7 \tiny{$\pm 0.2$} & 72.5 & 72.4 \tiny{$\pm 0.0$} & 40.8 \tiny{$\pm 0.2$}\\

TabICL \tiny{CAPGD}& 58.8 & 47.1 \tiny{$\pm 0.2$} & 20.8 \tiny{$\pm 0.2$} & 96.4 & 95.8 \tiny{$\pm 0.1$} & 31.0 \tiny{$\pm 0.4$} & 75.5 & 55.5 \tiny{$\pm 0.2$} & 50.3 \tiny{$\pm 0.2$} & 66.5 & 66.5 \tiny{$\pm 0.0$} & 10.8 \tiny{$\pm 0.2$} \\

\bottomrule
\end{tabular}}
\end{table*}

%% file: doc/appendices/proof.tex
\subsection{Proof Outline}\label{sec:app-proof}

In the following, we provide the proof of the convergence theorem under 5 assumptions, A1-A5,  detailed further below. 
We first recall the convergence theorem:

\begin{theorem}[Main Convergence Theorem]
Under the regularity and smoothness Assumptions  A1–A5, we have
\begin{align}
\lim_{t\to\infty}\lVert X'_{train}(t+1)-X'_{train}(t)\rVert & =0, \\
\lim_{t\to\infty}\lVert\nabla\widehat F_{val}(X'_{train}(t))\rVert & =0,
\end{align}
i.e., AICL iterations converge.
\end{theorem}

We recall that in AICL, one modifies the context $X'_{train} \in \mathbb{R}^{n\times d}$ while the model parameters $\theta$ remain fixed.

The associated robust optimization objective is given in Equation \ref{eq:aicl}. We note $F$ the associated expected adversarial loss, i.e.
\begin{multline}
    F(X') \defeq \\\mathbb{E}_{(X_{test},Y_{test}) \sim D}[\max_{\delta\in\Delta}
      \mathcal L\!\bigl(f(X_{test} + \delta\mid X',\,Y_{train},\theta),\,Y_{test}\bigr)].
\end{multline}

At iteration $t$ choose a single row $j(t)$ and replace it with: 

\begin{equation}
\label{eq:proof_2}
X'_{j(t)}(t+1)=
\Pi_{\mathcal C_{j(t)}}\!\bigl(
      X'_{j(t)}(t)+\eta_t\,d_{j(t)}(t)
\bigr).
\end{equation}

Essentially, this corresponds to one step of CAPGD attack. We make the following set of assumptions.

\begin{assumption2}[Context boundedness]
\textcolor{black}{The context perturbation $\delta^{\ast}$ is kept in the compact set $\Delta^*$.}
\end{assumption2}
\textcolor{black}{In our case, each foundation model has a fixed context size; feature values are clipped to known ranges. The projection step used in AICL enforces this bound exactly.}

\begin{assumption2}[Distribution match]

\textcolor{black}{The held-out validation set and the unseen test set follow the same distribution:  
$\operatorname{Dist}(X_{val},Y_{val})=\operatorname{Dist}(X_{test},Y_{test})$.}
\end{assumption2}

\textcolor{black}{In our case, standard tabular benchmarks are usually to be i.i.d. and split into train/test folds. The validation set is a subset of the train data that is never perturbed by the attack routine, so the equality of distributions is preserved throughout the algorithm.}

\begin{assumption2}[Model smoothness]

\textcolor{black}{There exists a finite constant $L>0$ such that the empirical robust loss}  

\textcolor{black}{
\begin{multline}
\widehat F_{val}(X')
\;=\; \\
\frac1m\sum_{(x,y)\in X_{val}}
      \max_{\delta\in\Delta}
      \mathcal L\!\bigl(f(x + \delta\mid X',\theta),\,y\bigr),
\end{multline}}

\end{assumption2} 
\textcolor{black}{has an $L$-Lipschitz gradient on the feasible set $\Delta$.}

\textcolor{black}{In our case, TabPFN v2 employs standard dot-product attention layers, which are Lipschitz on any bounded input domain if all intermediate operations are smooth, see \cite{pmlr-v139-kim21i}. However, we observed empirically that the network is sometimes numerically unstable. These non-smooth steps break the strict theoretical guarantee and can locally inflate the effective Lipschitz constant.  We acknowledge that Assumption 3 may fail in some cases; the convergence proof would then hold only up to the regions where the gradient remains well-behaved.}

\begin{assumption2}
Attack–replacement yields expected descent on held-out.

\textcolor{black}{Let $R_{j(t)}$ be the replacement operator that sets the $j(t)$-th row of $X'(t)$ to the adversarially perturbed version produced by the attack routine on an independent held-out mini-batch $(X_{val}^{(t)},Y_{val}^{(t)})$. There exists a constant $\alpha_0>0$ such that, for all $t$,
}

\textcolor{black}{\begin{multline}
\mathbb{E}\!\left[
  \widehat F_{\mathrm{val}}\!\bigl(R_{j(t)}(X'(t))\bigr)
  -
  \widehat F_{\mathrm{val}}\!\bigl(X'(t)\bigr)
  \;\middle|\; X'(t)
\right] \\
\le -\,\alpha_0\,
\bigl\| R_{j(t)}(X'(t))_{j(t)} - X'_{j(t)}(t) \bigr\|^{2}.
\tag{A4}
\end{multline}}
\end{assumption2}
\textcolor{black}{Equivalently, the one-row replacement by the attack produces a \emph{strict expected decrease} of the robust validation loss, with decrease at least quadratic in the replacement magnitude.}

\textcolor{black}{\emph{In our case}: The replacement $R_{j}$ is not a free direction but the adversarial output of a projected-gradient (PGD/FGSM-style) routine computed on an \emph{independent} held-out mini-batch, while the remainder of $X_{train}$ plays the role of context. Two ingredients make (A4) plausible:}

\begin{itemize}
\item \textcolor{black}{\textbf{Query–context coupling in Tabular FMs.} The model’s prediction depends on \emph{interactions between the query embedding} $q(x)$ (from the held-out batch) \emph{and the context key/value embeddings} $k(X'_{j})$, $v(X'_{j})$ via dot-product attention. Gradients of the robust loss with respect to the input $x$ and with respect to a context row $X'_{j}$ are therefore coupled through the same attention kernels and shared embedding stacks on a bounded domain. Consequently, an adversarial perturbation that \emph{increases} loss on queries induces, after sign flip in the replacement, a direction that \emph{decreases} the same loss with respect to the context row. Empirically we observe positive average cosine similarity between $-\nabla_{X'_j}\widehat F_{val}$ and the replacement increment $R_{j}(X')_j - X'_j$, providing a positive $\alpha_0$.} \\

\item \textcolor{black}{\textbf{Distribution matching via held-out swap.} At each iteration, the adversary is computed on a fresh held-out slice $X_{val}^{(t)}$ drawn i.i.d. from the test distribution (Assumption 2). Replacing one context row by its attacked counterpart makes the context better reflect the adversarially shifted test distribution considered in the inner max. This coupling ensures that—\emph{in expectation over the held-out mini-batch}—the robust validation loss does not increase, and decreases by an amount proportional to the replacement size.}
\end{itemize}

\textcolor{black}{We stress that (A4) is \emph{not} a gradient inner-product claim; it is a \emph{direct} descent property of the \emph{replacement operator} $R_j$ measured on an independent held-out mini-batch. In practice, if desired, (A4) can be enforced by a simple acceptance rule: compute the held-out robust loss before/after replacement; if it does not decrease, shrink $\eta_t$ and retry. This makes (A4) an operational condition rather than a latent property.}
\begin{assumption2}
\textcolor{black}{The context step sizes satisfy $\eta_t=\eta_0/(1+t)^\alpha$ with $0.5<\alpha\le1$.}
\end{assumption2}
\textcolor{black}{This is implied by using adaptive attack (e.g. CAPGD).\\}

\textcolor{black}{Define the constants:}  

\textcolor{black}{\begin{equation}
\label{eq:proof_3}
\eta_*=\frac{\alpha_0}{L},
\Hquad
\gamma_t=\bigl(\alpha_0-\tfrac{L}{2}\eta_t\bigr)\eta_t,
\Hquad
t\ge t_0\text{ with }\eta_t\le\eta_* .
\end{equation}}

\textcolor{black}{We can define and prove the following lemmas.}

\begin{lemma}

\textcolor{black}{The update scheme in Equation \ref{eq:proof_2} is continuous and graph-closed.}

\end{lemma}
\begin{proof}

Given by \textbf{Assumption 1} (bounded $\Delta^*$ ensures that the projection is continuous and $\Delta^*$ is compact.)\\

\end{proof}

\begin{lemma}

\textcolor{black}{$\nabla\widehat F_{val}$ is $L$-Lipschitz.}

\end{lemma}
\begin{proof}
Given by \textbf{Assumption 3}.
\end{proof}
\begin{lemma} We have, for all $t\geq 0$, 

\textcolor{black}{
\begin{multline}
\mathbb{E}\!\left[
  \widehat F_{\mathrm{val}}\!\bigl(X'(t{+}1)\bigr)
  \,\middle|\, X'(t)
\right]
\\
\le
\widehat F_{\mathrm{val}}\!\bigl(X'(t)\bigr)
\;-\;
\gamma_t\,
\bigl\| X'_{j(t)}(t{+}1)-X'_{j(t)}(t) \bigr\|^{2}.
\tag{L3}
\label{eq:suff-dec}
\end{multline}
}
    
\end{lemma}
\begin{proof}

Given \textbf{Assumption 3} (smoothness gives the quadratic upper bound), \textbf{Assumption 4} (direction correlates with $-\nabla$), and \textbf{Assumption 5} (step size small enough so $\gamma_t>0$)\\
\end{proof}
\begin{lemma}

\textcolor{black}{$\widehat F_{val}(X'(t))\downarrow F_\infty\!\ge\!0$}.
    
\end{lemma}
\begin{proof}
 Given by \textbf{Lemma 3} and \textbf{Assumption 5} ($\gamma_t>0$ and $\sum_t\eta_t^2<\infty$)\\
\end{proof}
\begin{lemma} 

\textcolor{black}{$\lVert\nabla_{j(t)}\widehat F_{val}(X'(t))\rVert\!\to\!0$}.

\end{lemma}
\begin{proof}
Given \textbf{Lemma 3} and \textbf{Assumption 5} (diverging $\sum_t\eta_t$ gives the summability of $\gamma_t$)
\end{proof}

\begin{lemma} Every limit point $X'^*$ satisfies $\nabla\widehat F_{val}(X'^*)=0$.
\end{lemma}
\begin{proof}
Given by \textbf{Lemma 5} + \textbf{Assumption 3} (gradient continuity) + \textbf{Assumption 1} (compactness guaranties the existence of cluster points)
\end{proof}

Finally, using these lemmas, we provide the proof of the convergence theorem stated above.

\begin{proof}[Proof of Convergence Theorem]
Every limit point $X'^*_{train}$ is first-order stationary for $\widehat F_{val}$. Because $\mathbb E[\widehat F_{val}(X')]=F(X')$ (\textbf{Assumption 2}), the same $X'^*_{train}$ is stationary for the true robust objective (Equation \ref{eq:aicl}) in expectation.  
\end{proof}

%% file: doc/appendices/budget_experiments_classical.tex
\subsection{\textcolor{black}{Additional Experimental Protocol Details}}\label{sec:app-exp-budget-classical}

\begin{figure*}[ht]
    \centering
    \begin{subfigure}[t]{0.49\textwidth}
        \centering
        \includegraphics[width=0.9\linewidth]{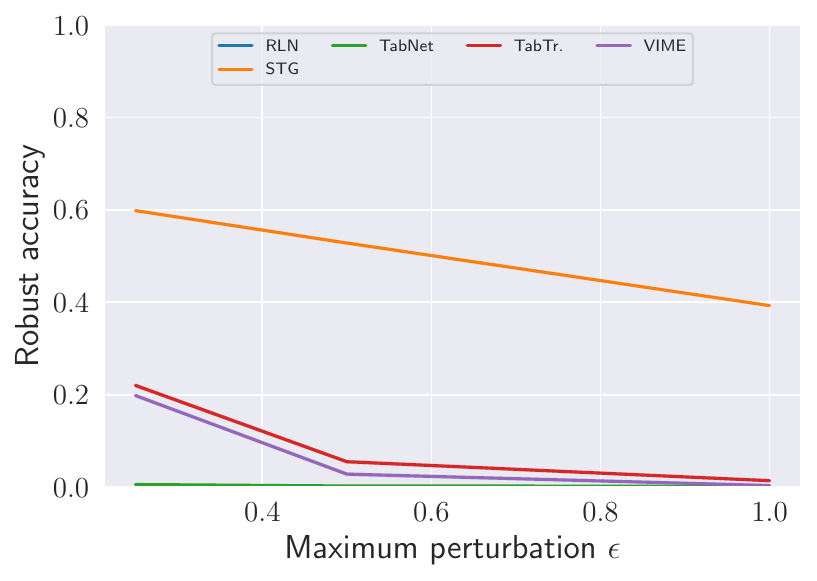}
        \caption{LCLD}
    \end{subfigure}%
    \hfill
    \begin{subfigure}[t]{0.49\textwidth}
        \centering
        \includegraphics[width=0.9\linewidth]{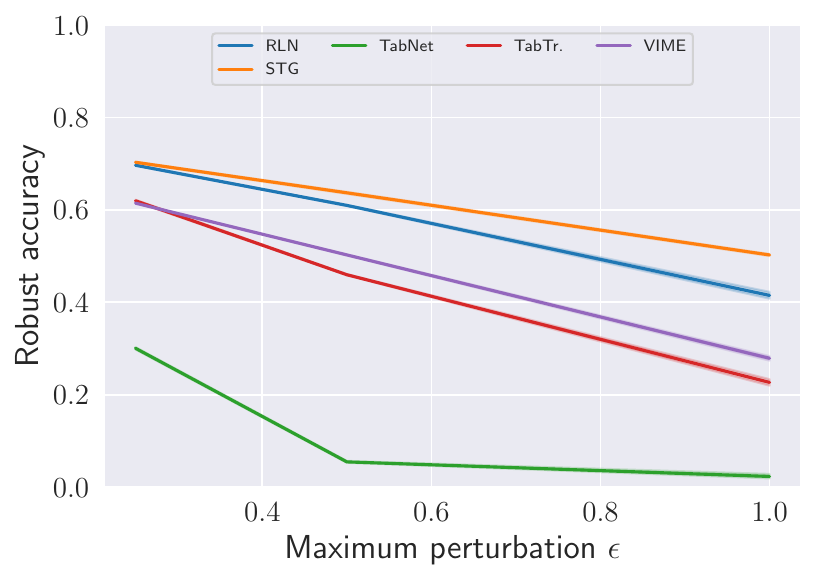}
        \caption{WIDS}
    \end{subfigure}
    \begin{subfigure}[t]{0.49\textwidth}
        \centering
        \includegraphics[width=0.9\linewidth]{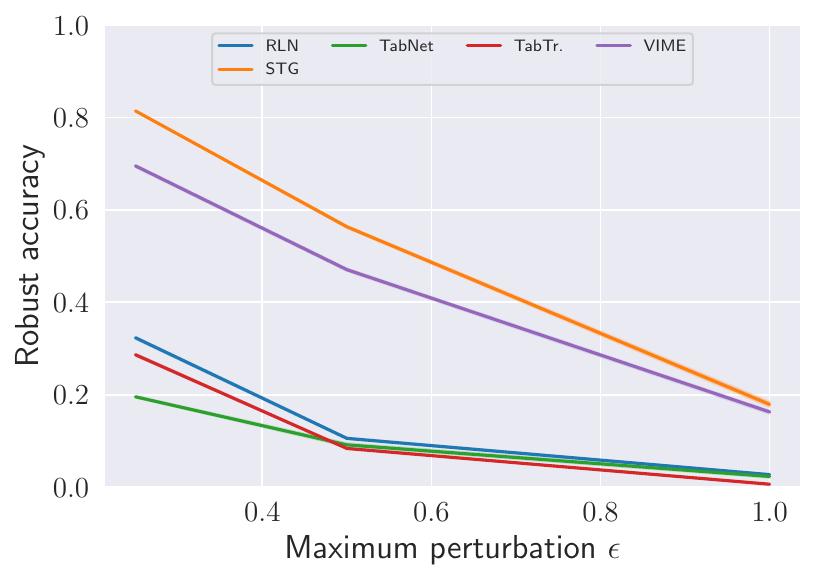}
        \caption{URL}
    \end{subfigure}
    \caption{Robust accuracy with CAA with varying maximum perturbation $\epsilon$.}
    \label{fig:classical-budget-eps}
\end{figure*}

\begin{figure*}[ht]
    \centering
    \begin{subfigure}[t]{0.49\textwidth}
        \centering
        \includegraphics[width=0.9\linewidth]{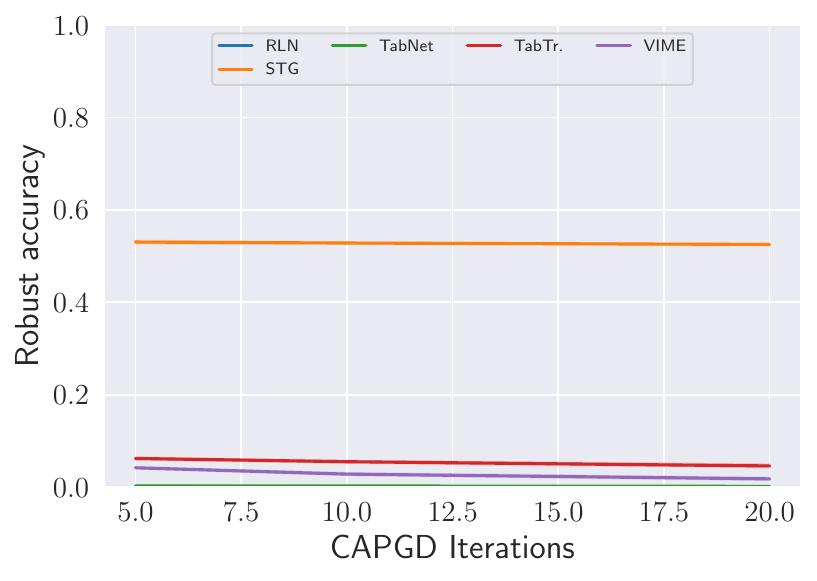}
        \caption{LCLD}
    \end{subfigure}%
    \hfill
    \begin{subfigure}[t]{0.49\textwidth}
        \centering
        \includegraphics[width=0.9\linewidth]{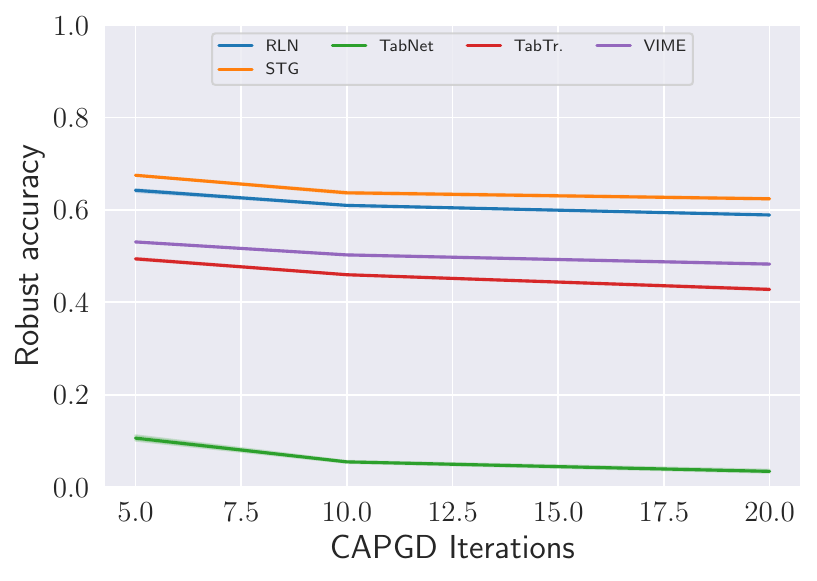}
        \caption{WIDS}
    \end{subfigure}
    \begin{subfigure}[t]{0.49\textwidth}
        \centering
        \includegraphics[width=0.9\linewidth]{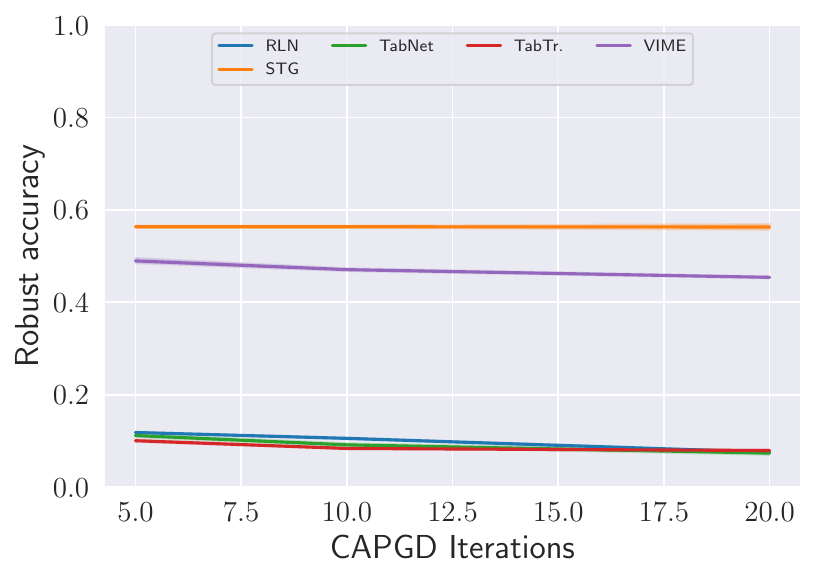}
        \caption{URL}
    \end{subfigure}
    \caption{Robust accuracy with CAA with varying gradient attack iterations in CAPGD.}
    \label{fig:classical-budget-steps}
\end{figure*}

We prolong the budget study on tabular non-FMs models. Figure \ref{fig:classical-budget-eps} shows $\epsilon$ budget variation, 
Figure \ref{fig:classical-budget-steps} for number of attack iterations of CAPGD (within CAA) and Figure \ref{fig:classical-budget-steps} for number of attack iterations of MOEVA (within CAA). Results are consistent with what was observed in previous research \cite{simonetto2024constrained}: $\epsilon$ drastically affects robustness of models, CAPGD number of iterations have little effect while MOEVA number of iterations have an impact mainly on URL dataset and on XGBoost and RandomForest. Note we did not include XGBoost and RandomForest in the first two figures as gradient attacks are not applicable.

Tabular-FMs follow similar trends to non-FMs model on $\epsilon$ budget and MOEVA number of iterations when comparing at fixed dataset. They however seems to be more affected by the number of iterations of CAPGD.

\begin{figure*}[ht]
    \centering
    \begin{subfigure}[t]{0.49\textwidth}
        \centering
        \includegraphics[width=0.9\linewidth]{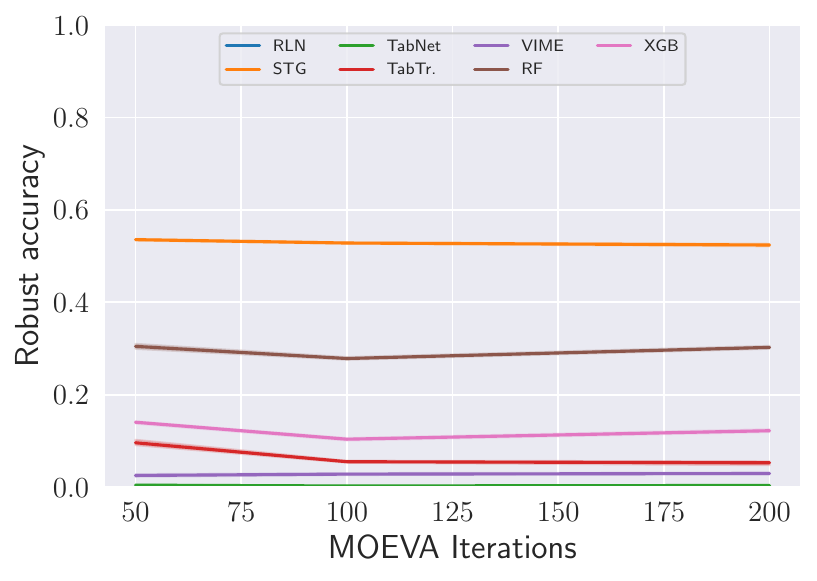}
        \caption{LCLD}
    \end{subfigure}%
    \hfill
    \begin{subfigure}[t]{0.49\textwidth}
        \centering
        \includegraphics[width=0.9\linewidth]{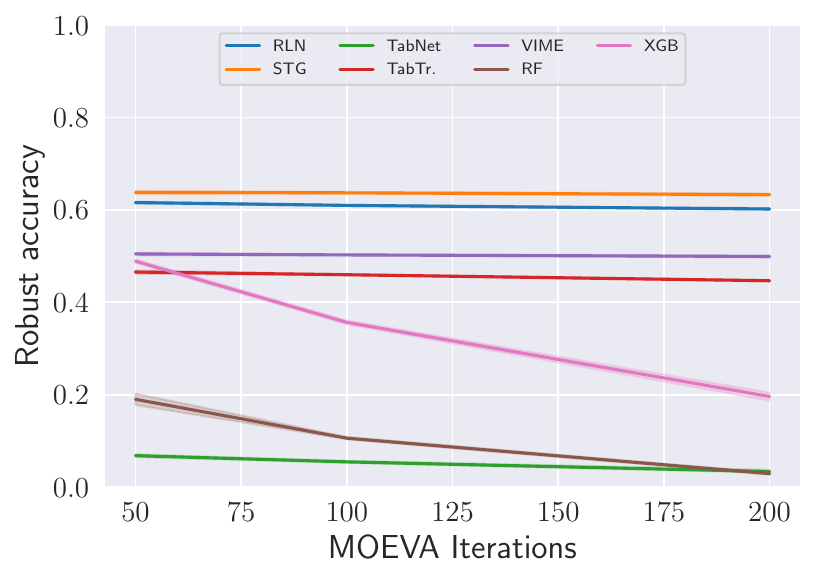}
        \caption{WIDS}
    \end{subfigure}
    \begin{subfigure}[t]{0.49\textwidth}
        \centering
        \includegraphics[width=0.9\linewidth]{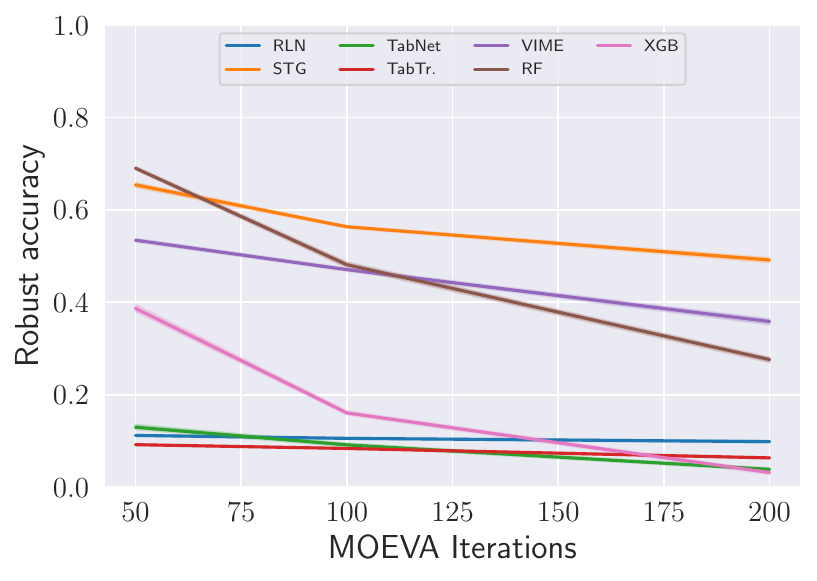}
        \caption{URL}
    \end{subfigure}
    \caption{Robust accuracy with CAA with varying search attack iterations in MOEVA.}
    \label{fig:classical-budget-n_gen}
\end{figure*}